\documentclass[nohyperref]{article}

\usepackage[accepted]{icml2022}

\usepackage[utf8]{inputenc} 
\usepackage[T1]{fontenc}
\usepackage{nicefrac}       
\usepackage{url}            

\usepackage{microtype}
\usepackage{graphicx}
\usepackage{subfigure}
\usepackage{booktabs} 

\usepackage{tabularx}

\usepackage{tikz}
\usetikzlibrary{bayesnet}

\usepackage{tcolorbox}

\usepackage[british]{babel}
\usepackage{natbib}

\usepackage{amsmath,amsthm,verbatim,amssymb,amsfonts,amscd}
\usepackage{thmtools,thm-restate}
\usepackage{mathtools}
\usepackage{mathrsfs}
\usepackage{commath}
\usepackage{bbm}

\usepackage{enumerate}
\usepackage{multirow}

\usepackage[colorlinks=true,
            linkcolor=black,
            citecolor=black]{hyperref}

\usepackage[capitalize,noabbrev]{cleveref} 

\definecolor{JungleGreen}{RGB}{41, 171, 135}
\definecolor{SkyBlue}{RGB}{78, 159, 229}
\definecolor{DarkRed}{RGB}{155, 0, 0}

\definecolor{yellowish}{RGB}{252, 232, 114}
\definecolor{salmon}{RGB}{234,153,153}
\definecolor{cornflowerblue}{RGB}{100,149,237}
\definecolor{paleblue}{RGB}{207,226,243}
\definecolor{taborange}{RGB}{222,143,6}
\definecolor{tabblue}{RGB}{76,156,201}

\allowdisplaybreaks

\theoremstyle{plain}
\newtheorem{theorem}{Theorem}[section]

\newtheorem{lemma}[theorem]{Lemma}

\theoremstyle{definition}

\theoremstyle{remark}

\DeclareMathOperator{\E}{\mathbb{E}}
\DeclareMathOperator{\Var}{\mathbb{V}}
\DeclareMathOperator{\Covar}{\mathbb{C}}
\DeclareMathOperator{\Prob}{\mathbb{P}}

\DeclareMathOperator{\given}{\vert}

\DeclareMathOperator{\diag}{diag}

\DeclarePairedDelimiterX{\infdivx}[2]{(}{)}{%
	#1 \, \delimsize\| \, #2%
}
\DeclarePairedDelimiterX{\commaparen}[2]{(}{)}{%
	#1 , #2%
}
\DeclarePairedDelimiterX{\innerprod}[2]{\langle}{\rangle}{%
	#1 , #2%
}
\DeclarePairedDelimiterX{\parenththree}[3]{\lparen}{\rparen}{
	#1 ; #2 ; #3	
}

\newcommand*\xbar[1]{%
	\hbox{%
		\vbox{%
			\hrule height 0.5pt 
			\kern0.5ex
			\hbox{%
				\kern-0.1em
				\ensuremath{#1}%
				\kern-0.1em
			}%
		}%
	}%
}

\newcommand{\KL}{\mathrm{KL}}
\newcommand{\TV}{\mathrm{TV} \commaparen}

\newcommand{\rv}[1]{\expandafter\MakeUppercase\expandafter{#1}}
\newcommand{\distr}[1]{\expandafter\MakeUppercase\expandafter{#1}}
\newcommand{\dens}[1]{#1}

\newcommand{\trace}{\mathrm{Tr}}
\newcommand{\natnum}{\mathbb{N}}

\newcommand{\indicatorSymbol}{\mathbbm{1}}
\newcommand{\indicator}[1]{\indicatorSymbol_{#1}}

\newcommand{\R}[1]{\mathbb{R}^{#1}}

\newcommand{\gauss}{\mathcal{N}}

\newcommand{\gp}{\mathrm{GP}}

\newcommand\restr[2]{{
  \left.\kern-\nulldelimiterspace 
  #1 
  \right|_{#2} 
  }}

\newcommand{\dataset}{\mathcal{D}}
\newcommand{\inputSpace}{\mathcal{X}}
\newcommand{\outputSpace}{\mathcal{Y}}

\newcommand{\layer}{\ell}
\newcommand{\nLayers}{L}
\newcommand{\nPoints}{n}
\newcommand{\dimParam}{d}
\newcommand{\minWidth}{\dimParam_{\min}}

\newcommand{\fn}{f}

\newcommand{\hidden}{h}
\newcommand{\nonlin}{\psi}

\newcommand{\weight}{W}
\newcommand{\bias}{b}
\newcommand{\wstd}{\sigma_\weight}
\newcommand{\bstd}{\sigma_\bias}
\newcommand{\lstd}{\sigma}
\newcommand{\kernelReg}{\lambda}
\newcommand{\posteriorMap}{T}
\newcommand{\altReparam}{R}

\newcommand{\param}{\theta}
\newcommand{\priorParam}{\phi}
\newcommand{\sparam}{\param^{\leq \nLayers}}
\newcommand{\spriorParam}{\priorParam^{\leq \nLayers}}
\newcommand{\sdimParam}{\dimParam^{\leq \nLayers}}
\newcommand{\rparam}{\param^{\nLayers + 1}}
\newcommand{\rpriorParam}{\priorParam^{\nLayers + 1}}
\newcommand{\rdimParam}{\dimParam^{\nLayers}}

\newcommand{\posterior}{\distr{P}_{\param \given \dataset }}
\newcommand{\rposterior}{\distr{P}_{\priorParam \given \dataset }}

\newcommand{\nngp}{K}
\newcommand{\enngp}{\Hat{\nngp}}
\newcommand{\regnngp}[1][\kernelReg]{\enngp_{#1}}
\newcommand{\embnngp}{\Psi}
\newcommand{\chonngp}{U}

\newcommand{\ntk}{\Theta}
\newcommand{\entk}{\Hat{\ntk}}
\newcommand{\regntk}[1][\kernelReg]{\entk_{#1}}

\newcommand{\residual}{r}
\newcommand{\weightedResidual}{\Tilde{\residual}}

\newcommand{\sg}{\mathtt{sg}}

\newcommand{\deq}{\mathrel{\mathop{:}}=}

\newcommand{\pp}[1]{(#1)}

\def \tabcellheight {1.3in}
\def \tabcellwidth {2.7in}
\newcommand{\centeredintabular}[1]{\begin{tabular}{l} #1 \end{tabular}}

\usepackage{scalerel}
\usepackage{nicefrac}
\usepackage{overpic}

\usetikzlibrary{arrows, arrows.meta}
\usetikzlibrary{positioning}
\usetikzlibrary{tikzmark}

\newcommand{\centeredintabularfixedwidth}[1]{%
\begin{minipage}{0.7in}%
\raggedright%
\noindent%
\unskip\parfillskip 0pt \par%
#1
\end{minipage}%
}

\newcommand{\tabhead}[1]{\textbf{ #1 }}
\newcommand{\tabinner}[1]{{%
\color{tabblue}#1}}

\icmltitlerunning{Wide Bayesian neural networks have a simple weight posterior: theory and accelerated sampling}

\begin{document}

\twocolumn[
\icmltitle{
Wide Bayesian neural networks have a simple weight posterior: \\theory and accelerated sampling
}




\begin{icmlauthorlist}
\icmlauthor{Jiri Hron}{goog,cam}
\icmlauthor{Roman Novak}{goog}
\icmlauthor{Jeffrey Pennington}{goog}
\icmlauthor{Jascha Sohl-Dickstein}{goog}
\end{icmlauthorlist}

\icmlaffiliation{cam}{University of Cambridge, UK}
\icmlaffiliation{goog}{Google Research, USA}

\icmlcorrespondingauthor{Jiri Hron}{jh2084@cam.ac.uk}

\icmlkeywords{Machine Learning, ICML}

\vskip 0.3in
]



\printAffiliationsAndNotice{}  


\begin{abstract}
    \looseness=-1
    We introduce \emph{repriorisation},
    a data-dependent reparameterisation which transforms a Bayesian neural network (BNN) posterior to a distribution whose KL divergence to the BNN prior vanishes as layer widths grow. 
    The repriorisation map acts directly on parameters, and its analytic simplicity 
    complements
    the known
    neural network Gaussian process (NNGP) behaviour of wide BNNs in
    function space.
    Exploiting the repriorisation,
    we develop a Markov chain Monte Carlo (MCMC) posterior sampling algorithm 
    which mixes \emph{faster} the wider the BNN. 
    This contrasts with the typically 
    poor performance of MCMC 
    in high dimensions.
    We observe up to 50x higher effective sample size relative to no reparametrisation for both fully-connected and residual networks.
    Improvements are achieved at all widths, with the margin between 
    reparametrised and standard
    BNNs growing with layer width.
\end{abstract}

\section{Introduction}

Bayesian neural networks (BNNs) have achieved significantly less success than their deterministic NN counterparts.
Despite recent progress \citep[e.g.,][]{khan18a, maddox2019, osawa2019, dusenberry20a, daxberger21a, izmailov2021bayesian}, 
(a)~our theoretical understanding of BNNs remains limited, and (b)~practical applicability is hindered by high computational demands.
We make progress on both these fronts.

Our work revolves around a reparametrisation $\param = \posteriorMap(\priorParam)$ of the (flattened) NN weights $\param \in \R{\dimParam}$.
Its form comes from a
theorem in which we establish that the Kullback--Leibler (KL) divergence between the standard normal distribution $\gauss(0, I_\dimParam)$ and the reparametrised \emph{posterior} $
\dens{p}(\phi \given \dataset)
= 
\dens{p}\pp{\posteriorMap(\priorParam) \given \dataset} 
\left| \det \partial_\priorParam \posteriorMap(\priorParam)\right|$
converges to zero as the BNN layers grow wide.\footnote{Case distinguishes distributions ($\distr{P}$) and their densities ($\dens{p}$). Densities are used in place of distributions where convenient.}
This closes a 
gap in the theory of overparametrised NNs by providing a rigorous characterisation of the BNN \emph{parameter space} behaviour (\Cref{fig:wide_net_research} and \Cref{sect:reparam}), and enables faster posterior sampling (\Cref{sect:sampling}).

\tcbset{width=2cm}

\begin{figure*}
    \centering
    \begin{tabular}{l @{\hspace{0.25em}}|@{\hspace{0.25em}}c@{\hspace{-1em}}c} 
         & \tabhead{parameter space} & \tabhead{function space} \\[0.2em]
         \hline \\[-0.5em]
        \centeredintabularfixedwidth{%
             \tabhead{Bayesian\\\ inference} 
        } & 
    \centeredintabular{ \begin{tcolorbox}[width=\tabcellwidth,height=\tabcellheight,colback=tabblue!75,boxrule=0pt,colframe=white,coltext=white,arc=0pt,outer arc=0pt]
            \centering\footnotesize
            \tabinner{\color{white} \bf 
            {Repriorisation}: posterior $\rightarrow$ prior}  \\
             {\tiny \hypersetup{linkcolor=white}
             \Cref{eq:full_reparam}, \Cref{stm:zero_kl} (this paper)}\\[-1.1em]        
            {\begin{align*}
                &\priorParam^{\layer}
                \coloneqq
                \begin{cases}
                    \,\Sigma^{-1/2} (\param^{\layer} - \mu )
                    & 
                    \qquad \layer = \nLayers + 1,
                    \\
                    \,\param^{\layer}
                    &
                     \qquad \text{otherwise.}
                \end{cases}
                \\[1em]
                &\KL\pp{\, \gauss(0, I_{\dimParam}) \, \| \, \rposterior \,} \to 0 
                \text{ as } \minWidth \to \infty
             \end{align*}}
        \end{tcolorbox} } &
        \centeredintabular{ \begin{tcolorbox}[width=\tabcellwidth,height=\tabcellheight,boxrule=0pt,colframe=white,colback=gray!8,arc=0pt,outer arc=0pt]    
            \centering\footnotesize
            \tabinner{\bf Neural Network Gaussian Process (NNGP)} \\
             {\tiny \citep{matthews2018gaussian,lee2018deep,hron2020exact}} \\[-0.5em]        
            {
                \begin{gather*}
                    \theta \sim \distr{P}_\theta
                    \, \, 
                    \overset{\minWidth}{\longrightarrow}
                    \, \,
                    f_{\theta} \sim \gp (0, k)
                    \\
                    \theta \sim \distr{P}_{\theta \given \dataset}
                    \, \, 
                    \overset{\minWidth}{\longrightarrow}
                    \, \,
                    f_{\theta} \sim \gp (m_{k; \dataset}, S_{k; \dataset})
                \end{gather*}
            }
        \end{tcolorbox} } \\ 
        \centeredintabularfixedwidth{%
             \tabhead{gradient\\\ descent\\\ training} 
        } & 
        \centeredintabular{ \begin{tcolorbox}[width=\tabcellwidth,height=\tabcellheight,boxrule=0pt,colframe=white,colback=gray!8,arc=0pt,outer arc=0pt] 
            \centering\footnotesize
            \tabinner{\bf Linearisation} \\
            {\tiny \citep{lee2019wide,chizat2019lazy}}\\[-0.5em] 
             {
             \begin{gather*}
                 \partial_t \theta_t
                 =
                 -\eta\, \nabla_{\theta_t} \,\mathcal{L}\pp{y, \fn_{\theta_t}(X)} 
                 \\[1em]
                \fn_{t}\pp{x} = 
                    \fn_{0}\pp{x} + 
                    \tfrac{
                    \partial f_{0} (x)
                    }{
                    \partial \theta_0}
                    \pp{\theta_t - \theta_0} 
                    +
                    \mathcal{O}\pp{
                        \minWidth^{-1/2} 
                 }
             \end{gather*}
             }
        \end{tcolorbox} 
        } &
        \centeredintabular{ \begin{tcolorbox}[width=\tabcellwidth,height=\tabcellheight,boxrule=0pt,colframe=white,colback=gray!8,arc=0pt,outer arc=0pt]    
            \centering\footnotesize
            \tabinner{\bf Neural Tangent Kernel (NTK)} \\
            {\tiny \citep{jacot2018neural,allen2019convergence,du2019gradient}}\\[-0.5em]        
             {
             \begin{gather*}
                 \partial_t \fn_t(x) 
                 =
                 -\eta\, \hat{\Theta}_{x, X} \nabla_{\fn_t(X)} \,\mathcal{L}\pp{y, \fn_t(X)} 
                 \\[1em]
                 \fn_t \mid \dataset
                 \sim 
                 \gp\pp{
                    m_{t; \dataset}, S_{t; D}
                 }
                \text{ as } \minWidth \to \infty
             \end{gather*}
             }
        \end{tcolorbox} } 
    \end{tabular}
    \caption{\textbf{As neural networks are made wide, their behaviour often becomes simple.} 
    Past results 
    examine wide NNs in either function space or parameter space, and 
    either under gradient descent training or Bayesian inference. The central result for each condition is stated, 
    using this paper's formalism where applicable.
    The behaviour of wide Bayesian NNs in parameter space is largely unexplored. 
    We address this gap,
    by reparametrising the weight posterior so that its KL divergence from the prior $\gauss (0, I_{\dimParam})$ vanishes with increasing width.
    }
    \label{fig:wide_net_research}
\end{figure*}

In \emph{function space}, wide BNN \emph{priors} converge (weakly) to a so-called neural network Gaussian process (NNGP) limit \citep{matthews2018gaussian,lee2018deep,alonso2018deep,novak2019bayesian,yang2019scaling,hron2020infinite}, 
and the \emph{posteriors} converge (weakly) to that of the corresponding NNGP \citep[assuming the likelihood is a bounded continuous function of the NN outputs;][]{hron2020exact}.
\emph{Parameter space} behaviour is less understood \citep{hron2020exact}.
The main exception is the work of \citet{matthews2017sample}
who showed that randomly initialising a NN, and then optimising only the last layer with respect to a mean squared error loss, is equivalent to drawing a sample from the \emph{conditional last layer posterior}.
This result holds for a Gaussian likelihood and zero observation noise.
Our reparametrisation $\posteriorMap$ can be seen as a non-zero noise generalisation of the underlying map from the initial to the optimised weights, and provides samples from the \emph{joint posterior} over all layers in the infinite width limit (\Cref{stm:zero_kl}).

Since sampling from the `KL-limit' $\gauss(0 , I)$ is trivial relative to the notoriously complex BNN posterior, our theory motivates applying Markov chain Monte Carlo (MCMC) to the reparametrised density.
To apply MCMC, two concerns must be addressed: 
(i)~understanding if the KL-closeness to $\gauss(0 , I_\dimParam)$ actually simplifies sampling, and
(ii)~computational efficiency.
In \Cref{sect:gradients}, we partially answer (i) by proving that 
the gradient $\nabla_{\priorParam} \log \dens{p} (\priorParam \given \dataset)$
concentrates around the log density gradient of $\gauss ( 0 , I)$ in wide BNNs.
For (ii), we propose a simple-to-implement and 
efficient way of computing both $\posteriorMap(\priorParam)$ and the corresponding density at the same time using Cholesky decomposition (\Cref{sect:cholesky}).

Applying Langevin Monte Carlo (LMC), we empirically demonstrate up to 50x improved mixing speed, as measured by effective sample size (ESS).
The phenomenon occurs both for residual and fully-connected networks (FCNs), and for a variety of dataset sizes and hyperparameter configurations.
The improvement over no reparametrisation increases with layer width, but 
is observed
even far from the NNGP regime.
For example,
we find a 10x improvement on \texttt{cifar-10} for a 3-hidden layer FCN with 1024 units per layer.
However, for ResNet-20, a 10x improvement occurs
only when the top-layer width $\rdimParam$ is similar to or larger than the number of observations, illustrating that gains are possible but not guaranteed outside of the NNGP regime.

\subsection{Assumptions and notation}
\label{sect:notation_assumptions}

A BNN models a mapping from inputs $x \in \inputSpace$ to outputs $y \in \outputSpace$ using a parametric function $\fn \coloneqq  \fn_\param$.
An example is an $\nLayers$-hidden layer fully-connected network $\fn_\param = \fn^{\nLayers + 1}$ with
\begin{align}\label{eq:fcn}
    \fn^{\layer} (x)
    =
    \tfrac{\wstd^{\layer}}{\sqrt{\dimParam^{\layer-1}}}
    \hidden^{\layer - 1}(x) \weight^{\layer} 
    \, ,
    \quad
    \hidden^{\layer}(x) = \nonlin(\fn^{\layer}(x))
    \, ,
\end{align}
with $\nonlin$ the nonlinearity, $\hidden^0 (x) \deq x$, and $\weight^{\layer} \in \R{\dimParam^{\layer-1} \times \dimParam^{\layer}}$
(bias terms are wrapped into $\weight^{\layer}$ by adding a constant entry dimension to $\hidden^{\layer - 1}(x)$).
$\param^{\layer}$ will denote the flattened $\layer$\textsuperscript{th} layer parameters,
and $\param \coloneqq [ \param^\layer]_{\layer=1}^{\nLayers + 1} \in \R{\dimParam}$ their concatenation.
Later on, the vector of readout weights $\rparam$ will be of special importance, as will the \textbf{minimum hidden layer width} $\boldsymbol{\minWidth \coloneqq \min_{1 \leq \layer \leq \nLayers} \dimParam^{\layer}}$.
While we will study how the behaviour of $\fn$ and $\param$ changes with layer width, we suppress this dependence in our notation to reduce clutter.

The factor
$\nicefrac{\wstd^{\layer}}{\sqrt{\dimParam^{\layer-1}}}$ 
in \Cref{eq:fcn} is more commonly part of the weight prior.
We use this so-called `NTK parametrisation' as it allows taking $\gauss(0, I)$ as the prior regardless of the network width, which simplifies our notation without changing the implied function space distribution.
Our claims hold under the standard parametrisation as well, by making
multiplication by 
$\nicefrac{\wstd^{\layer}}{\sqrt{\dimParam^{\layer-1}}}$
a part of the reparametrisation.

We assume the final readout layer is linear, and that the likelihood is Gaussian $\dens{p}(y \given X, \param) \propto \exp \{ -\tfrac{1}{2 \lstd^2}\sum_{i=1}^\nPoints (y_i - f_{\param}(x_i))^2 \}$ with observation variance $\lstd^2 > 0$.
In contrast, the rest of the network need not be an FCN, but can contain any layers for which NNGP behaviour is known, including convolutional and multi-head attention layers, and skip connections \citep[see, e.g.,][for an overview]{yang2020tensor}.

\section{Repriorisation: \texorpdfstring{posterior $\to$ prior}{mapping posterior to prior}}
\label{sect:reparam}

\subsection{Repriorisation of linear models}
\label{sect:lin_models}

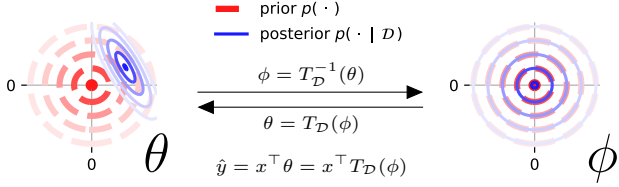
\begin{figure}[tbp]
    \centering
    \begin{tabular}[t]{@{\hspace{0em}}c @{\hspace{1.4em}} c @{\hspace{0.9em}} c}
        \begin{overpic}[width=0.8in,trim=7 8 6 6]{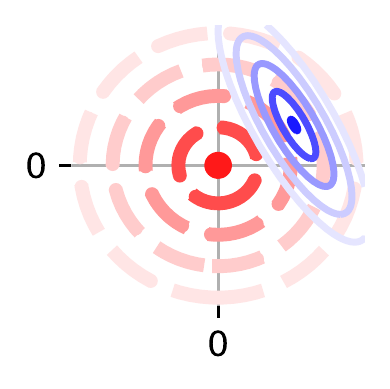}
            \put (90,0) {\huge $\theta$}
        \end{overpic}    
        & 
              {\begin{minipage}[b]{1.2in}%
                    \centering%
                    \noindent%
                    \unskip\parfillskip 0pt \par%
            \begin{overpic}[width=1in,trim=3 7 3 2]{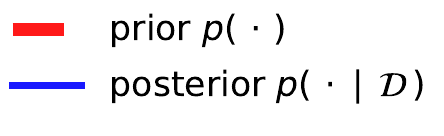}
            \end{overpic}  \\[0.5em]  
             \scriptsize 

                 $\priorParam = T_\dataset^{-1}\pp{\theta }$ \\[0.15em]
                    \begin{tikzpicture}[scale=1, transform shape]
                        \draw[>=triangle 45, <-] (0,0) -- (3,0);
                        \draw[>=triangle 45, ->] (0,0.2) -- (3,0.2);
                    \end{tikzpicture}
                    $\theta = T_\dataset\pp{\priorParam}$
                    \\[1em]
                 $\hat{y} = x^\top \theta = x^\top T_\dataset \pp{\phi}$
\end{minipage}}%
        &
        \begin{overpic}[width=0.8in,trim=7 8 6 6]{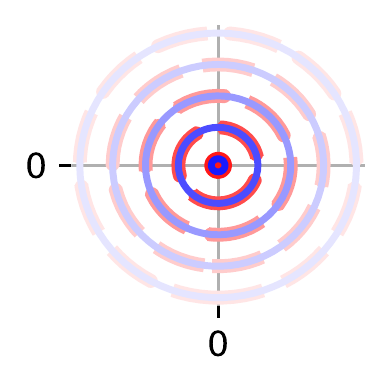}
            \put (90,0) {\huge $\phi$}
        \end{overpic}    
    \end{tabular}
    \caption{
    \textbf{Reparameterisation of Bayesian \emph{linear} regression maps the posterior to the prior.}
    \textbf{Left:} Shows the parameter $\param$ prior, and the posterior for a single datapoint $X = [ 0.9, 0.5 ], \ y = [ 2 ]$, with $\lstd^2 = 0.1$. 
    \textbf{Right:} After a data-dependent affine reparameterisation, described in \Cref{sect:lin_models},
    the posterior becomes identical to the prior in terms of new parameters $\priorParam$. 
    MCMC sampling from the parameter posterior is easier in $\priorParam$ than in $\param$.
    A similar reparameterisation can be applied to deep BNNs.}
    \label{fig:lin_reparam}
\end{figure}

To begin,
we consider a Bayesian linear model 
$y \given \param , x \sim \gauss (x^\top \param, \lstd^2)$.
For a standard normal prior $\param \sim \gauss(0, I_\dimParam)$,
the Bayesian posterior after observing $\nPoints$ points, $\dataset \coloneqq (X, y) \in \R{\nPoints \times \dimParam} \times \R{\nPoints}$, is a Gaussian $\param \given \dataset \sim \gauss (\mu , \Sigma)$ with
\begin{align}\label{eq:lin_posterior_params}
    \Sigma 
    =
    (I_\dimParam + \lstd^{-2}X^\top X)^{-1}
     , \, \text{and }
    \mu 
    = 
    \lstd^{-2} \Sigma X^\top y
    \, .
\end{align}
We can thus \emph{transform a posterior into a prior sample} using the data-dependent reparameterisation $\priorParam = T^{-1}\pp{\theta} = \Sigma^{-1/2}(\param - \mu) \sim \gauss(0, I_\dimParam)$ as illustrated in \Cref{fig:lin_reparam}.

\subsection{Repriorisation of Bayesian neural networks}
\label{sect:reparam_nonlinear}

A similar insight applies to BNNs with the assumed Gaussian prior and likelihood.
Specifically,
the posterior of readout (top-layer) weights \emph{conditioned} on the other parameters is $\rparam \given \sparam, \dataset \sim \gauss(\mu, \Sigma)$ with $\mu$ and $\Sigma$ as in \Cref{eq:lin_posterior_params}, except $X$ is replaced by the scaled top-layer embeddings 
$
\embnngp \coloneqq 
\nicefrac{ 
    \wstd^{\nLayers + 1}
    \hidden^{\nLayers}(X)
}{
    \sqrt{\rdimParam}
} 
\in \R{\nPoints \times \rdimParam}
\!$.
Our reparametrisation $\priorParam = \posteriorMap^{-1}(\param)$ can thus be defined analogously:%
\footnote{
See \Cref{app:alt_reparam} for discussion of alternative reparametrisations which modify parameter values in layers beyond readout.
}
\begin{align}\label{eq:full_reparam}
    \priorParam^{\layer}
    =
    \begin{cases}
        \Sigma^{-1/2} (\param^{\layer} - \mu )
        & 
         \qquad \layer = \nLayers + 1,
        \\
        \param^{\layer}
        &
         \qquad \text{otherwise.}
    \end{cases}
\end{align}
This ensures $\rpriorParam \given \spriorParam, \dataset \sim \gauss(0, I_{\rdimParam} )$ for any fixed value of $\spriorParam = \sparam$.
We omit the dependence of $\mu$, $\Sigma$ on the dataset $\dataset$ and the pre-readout parameters $\sparam$ to reduce clutter, but emphasise that $\posteriorMap$ depends on both.

Since $\posteriorMap$ is a differentiable bijection, the full reparametrised density is $\dens{p} (\priorParam \given \dataset) = \dens{p} (\param \given \dataset) \left|  \det \partial_\priorParam \param \right|$
where
\begin{align}\label{eq:determinant}
    \det (\partial_\priorParam \param)
    =
    \det
    \begin{pmatrix}
        \Sigma^{1/2} &
        \tfrac{\partial \rparam}{\partial \spriorParam}
        \\
        0 &
        I_{(\dimParam - \rdimParam)}
    \end{pmatrix}
    =
    \sqrt{\det(\Sigma)}
    \, .
\end{align}

\subsection{Interpreting the reparametrised distribution}

To better understand the reparametrisation, note 
$\dens{p} (\priorParam \given \dataset) = \dens{p}(\rpriorParam \given \spriorParam, \dataset) \, \dens{p}(\spriorParam \given \dataset)$
where we already know the first term is equivalent to $\gauss(0 , I_{\rdimParam})$.
Since $\spriorParam = \sparam$, the latter term is equal to the \emph{marginal} posterior over $\sparam$
\begin{align}\label{eq:remainder_posterior}
    &\dens{p}(\spriorParam \given \dataset)
    \propto
    \dens{p} (\sparam)
    \, \E_{\rparam \sim \gauss(0, I_{\rdimParam})} \left[
        \dens{p} (y \given \param, X) 
    \right]
    \nonumber
    \\
    &\overset{\text{(i)}}{\propto}
    \, \dens{p}(\sparam)
    \, \sqrt{\det \Sigma}
    \, \exp \bigl\{
        \tfrac{1}{2} \,
        y^\top (\lstd^2 I_\nPoints + \embnngp \embnngp^\top)^{-1} y
    \bigr\}
\end{align}
where
(i)~comes from completing the square $\| \rparam - \mu \|_{\Sigma^{-1}}^2$ ($\|v\|_A^2 \coloneqq v^\top A v$),
and applying the Woodbury identity.

Using the Weinstein--Aronszajn identity
\begin{align*}
    \det (\Sigma)
    =
    \det (I_{\rdimParam} + \lstd^{-2} \embnngp^\top \embnngp)
    \propto
    \det (\lstd^2 I_{\nPoints} + \embnngp \embnngp^\top)
    \, ,
\end{align*}
the readers familiar with the GP literature may recognise $\dens{p}(\spriorParam \given \dataset)$ as $\dens{p}(\sparam)$ weighted by the marginal likelihood of a centred GP with the \emph{empirical NNGP kernel} $\regnngp[\lstd^2] \coloneqq \lstd^2 I_\nPoints + \embnngp \embnngp^\top$ \citep{rasmussen2005gps}.
The marginal likelihood is often used for optimisation of kernel hyperparameters.
Here the role of the `hyperparameters' is played by $\sparam$, i.e., all but the readout weights.

This relation to the empirical NNGP kernel is crucial in the next section where we exploit the known convergence of $\regnngp[\lstd^2]$ to a constant \emph{independent} of $\sparam$ in wide BNNs, to prove that the reparametrised posterior converges to the prior distribution at large width.
\Cref{fig:posterior_to_prior} provides an informal argument motivating that same convergence, using the language of probabilistic graphical models (PGMs).

\begin{figure}[tbp]
    \centering
    \includegraphics[keepaspectratio,width=0.74\linewidth]{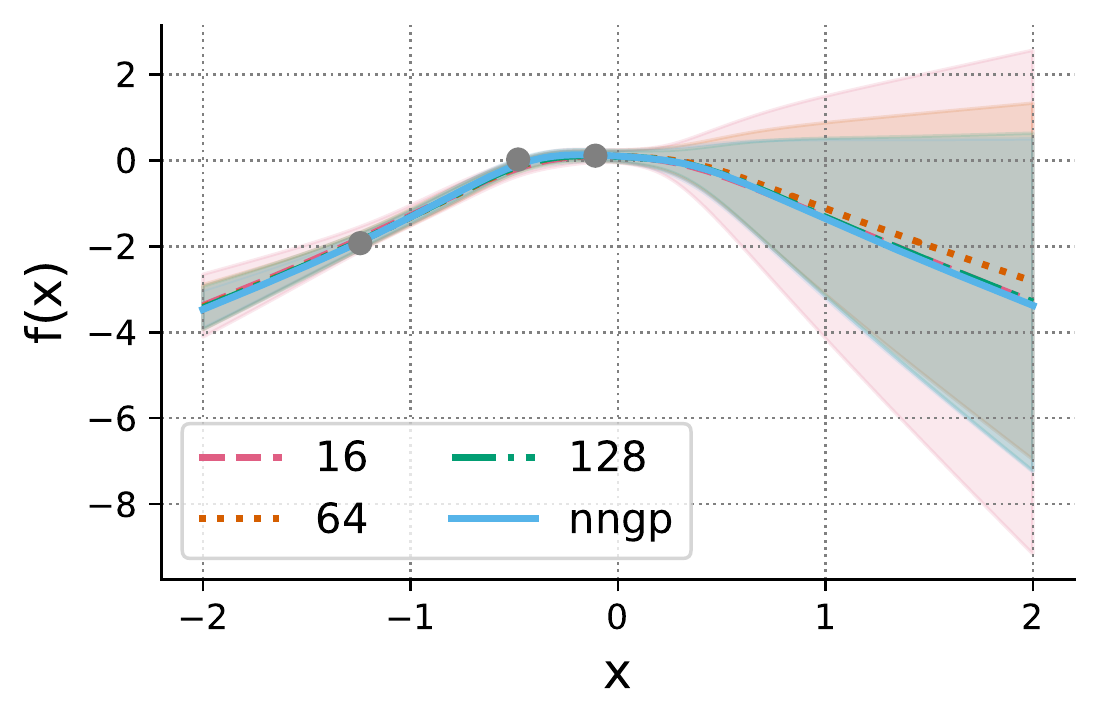}
    \vspace*{-1em}
    \caption{
        \textbf{The functions parametrised by $\posteriorMap(\priorParam)$, $\priorParam \sim \gauss(0, I)$, converge to the true posterior.} 
        The distribution over functions $\fn_{\posteriorMap(\priorParam)}$ 
        is shown for a 3-hidden layer FCN with 1D inputs and outputs, and 3 datapoints (dark circles). 
        As layer width increases (legend), the BNN posterior converges to the NNGP posterior \citep{hron2020exact}, and so does $\fn_{\posteriorMap(\priorParam)}$ (our \Cref{stm:kl_functions}).
    }
    \label{fig:reparam_convergence}
    \vspace{-0.625\baselineskip}
\end{figure}

\subsection{Asymptotic normality under reparametrisation}
\label{sect:asympt_normality}

\newcommand{\pgmbase}{0.5cm}
\begin{figure*}
\centering
\begin{tabularx}{\linewidth}{ c | c | c | c }
& & & \\  
\qquad
\begin{tikzpicture}
  \node[obs]    at (-0.8, -1)       (x) {$X$};
  \node[latent, dashed, above=\pgmbase of x] (hL) {$h^L$};
  \node[latent, dashed, above=\pgmbase of hL] (hLp) {$f^{L+1}$};
  \node[obs, above=\pgmbase of hLp] (y) {$y$};  
  \node[latent, left=\pgmbase of hL] (phiL) {$\spriorParam$};  
  \node[latent, left=\pgmbase of hLp] (phiLp) {$\rpriorParam$};  

  \edge {x,phiL} {hL} ; %
  \edge {hL,phiLp} {hLp} ; %
  \edge {hLp} {y} ; %
  
  \node[left=of x] (label) {\small\textbf{(a)}};
\end{tikzpicture}
\tikzmark{a2}
\qquad\qquad
& 
\qquad
\tikzmark{b1}
\begin{tikzpicture}
  \node[obs] at (3, 2)   (x) {$X$};
  \node[latent, dashed, above=\pgmbase of x] (hL) {$h^L$};
  \node[obs, above=\pgmbase of hL] (y) {$y$};  
  \node[latent, left=\pgmbase of hL] (phiL) {$\spriorParam$};  

  \edge {x,phiL} {hL} ; %
  \edge {hL} {y} ; %
  
  \node[left=of x] (label) {\small\textbf{(b)}};
\end{tikzpicture}
\tikzmark{b2}
\qquad\qquad
& 
\qquad
\tikzmark{c1}
\begin{tikzpicture}

  \node[obs] at (3, 2)   (x) {$X$};
  \node[latent, dashed, above=\pgmbase of x] (KL) {$\enngp^L$};
  \node[obs, above=\pgmbase of KL] (y) {$y$};  
  \node[latent, left=\pgmbase of KL] (phiL) {$\spriorParam$};  

  \edge {x,phiL} {KL} ; %
  \edge {KL} {y} ; %
  
  \node[left=of x] (label) {\small\textbf{(c)}};
\end{tikzpicture}
\tikzmark{c2}
\qquad\qquad
& 
\qquad
\tikzmark{d1}
\begin{tikzpicture}

  \node[obs]     at (3, -2.3)           (x) {$X$};
  \node[latent, dashed, above=\pgmbase of x] (KL) {$\enngp^L$};
  \node[obs, above=\pgmbase of KL] (y) {$y$};  
  \node[latent, left=\pgmbase of KL] (phiL) {$\spriorParam$};  

  \edge {x} {KL} ; %
  \edge {KL} {y} ; %
  
  \node[left=of x] (label) {\small\textbf{(d)}};
\end{tikzpicture}
\qquad\qquad
\\[0.2cm]
\end{tabularx}

\def\arrowoffsety{3.8 cm}
\def\arrowoffsetx{0.3 cm}
\def\boxoffsety{1.3 cm}
\def\boxoffsetyy{-0.3 cm}
\def\textoffsety{0.15 cm}
\begin{tikzpicture}[overlay, remember picture]

\fill [white] ([xshift=\arrowoffsetx,yshift=\arrowoffsety+\boxoffsety]pic cs:a2) rectangle ([xshift=\arrowoffsetx,yshift=\arrowoffsety+\boxoffsetyy]pic cs:b1);
\draw[->,blue,very thick,color=tabblue] ([xshift=\arrowoffsetx,yshift=\arrowoffsety]pic cs:a2) -- ([xshift=\arrowoffsetx,yshift=\arrowoffsety]pic cs:b1) node [above,midway,yshift=\textoffsety,color=tabblue] {\shortstack{marginalise\\top layer}};

\fill [white] ([xshift=\arrowoffsetx,yshift=\arrowoffsety+\boxoffsety]pic cs:b2) rectangle ([xshift=\arrowoffsetx,yshift=\arrowoffsety+\boxoffsetyy]pic cs:c1);
\draw[->,blue,very thick,color=tabblue] ([xshift=\arrowoffsetx,yshift=\arrowoffsety]pic cs:b2) -- ([xshift=\arrowoffsetx,yshift=\arrowoffsety]pic cs:c1) node [above,midway,yshift=\textoffsety,color=tabblue] {\shortstack{$h^L$ only influences $y$\\through kernel $\enngp^L$}};

\fill [white] ([xshift=\arrowoffsetx,yshift=\arrowoffsety+\boxoffsety]pic cs:c2) rectangle ([xshift=\arrowoffsetx,yshift=\arrowoffsety+\boxoffsetyy]pic cs:d1);
\draw[->,blue,very thick,color=tabblue] ([xshift=\arrowoffsetx,yshift=\arrowoffsety]pic cs:c2) -- ([xshift=\arrowoffsetx,yshift=\arrowoffsety]pic cs:d1) node [above,midway,yshift=\textoffsety,color=tabblue] {\shortstack{$\enngp^L \rightarrow $ constant\\as $\minWidth \rightarrow \infty$}};
\end{tikzpicture}
\caption{
\textbf{
    Sketch motivating 
    convergence 
    of the reparametrised posterior
    to the prior.
}
\textbf{(a)}~A probabilistic graphical model (PGM) describing the BNN \emph{after reparameterisation}.
Shaded circles correspond to observed variables. 
Dashed outlines indicate that the variable value is a deterministic function of its ancestors. 
The reparametrised posterior factorises as $p(\priorParam \given \dataset) = \dens{p}(\rpriorParam \given \spriorParam, \dataset)\, \dens{p}( \spriorParam \given \dataset)$. 
\textbf{(b)}~Our reparametrisation in \Cref{eq:full_reparam} makes $\rpriorParam \given \spriorParam, \dataset \sim \gauss(0, I)$ for any $\spriorParam$.
The marginal $\dens{p}( \spriorParam \given \dataset) = \int \! \dens{p}( \priorParam \given \dataset) \, \textrm{d} \rpriorParam$ can be evaluated analytically with a Gaussian likelihood (\Cref{eq:remainder_posterior}).
\textbf{(c)}
$\dens{p}(\spriorParam \given \dataset)$ depends on the data only via
the (empirical) NNGP kernel 
$
\enngp^L
= 
\hidden^{\nLayers}(X) 
\hidden^{\nLayers}(X)^\top
/ \rdimParam
$.
This is due to the assumed linearity of the top layer, the Gaussian prior, and the identity $A \varepsilon \sim \gauss (0, A A^\top)$ for $\varepsilon \sim \gauss (0, I)$ and any fixed matrix $A$.
\textbf{(d)}~As layer width $\minWidth$ goes to infinity, 
$\enngp^L$
converges to a constant independent of $\spriorParam$
(\Cref{sect:asympt_normality}). 
Hence $\dens{p}( \spriorParam \given \dataset) \approx \dens{p}( \spriorParam ) = \gauss(0, I)$ by the PGM. 
Since both $\rpriorParam$ and $\spriorParam$ are approximately Gaussian, and $\rpriorParam$ is independent of $\spriorParam$, the joint posterior is also approximately Gaussian
$\dens{p}(\priorParam \given \dataset) = \dens{p}(\rpriorParam \given \spriorParam, \dataset) \dens{p}( \spriorParam \given \dataset) \approx \gauss(0, I)$.
}
\label{fig:posterior_to_prior}
\end{figure*}
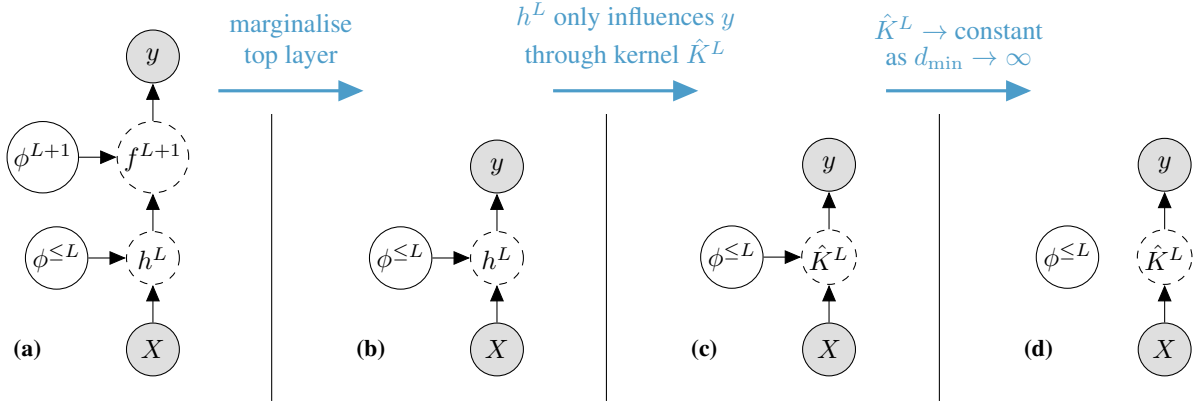

Our reparametrisation guarantees normality of the reparametrised posterior over $\rpriorParam$ for \emph{any} width.
\Cref{eq:remainder_posterior} suggests 
$
\spriorParam
$
may exhibit Gaussian behaviour as well in wide BNNs, since---outside of $\dens{p}(\sparam)$---its posterior depends on $\spriorParam = \sparam$ only through the 
empirical NNGP kernel
which is known to converge a constant $\nngp \in \R{\nPoints \times \nPoints}$ as the layer width grows \citep[see][for an overview]{yang2020tensor}.

This is crucial in establishing our first major result: convergence of the $\KL$ divergence between the $\gauss(0, I_\dimParam)$ prior and the reparametrised posterior to zero as the number of hidden units in each layer goes to infinity.\footnote{Please refer to \Cref{app:proofs} for all proofs.}
\begin{restatable}{theorem}{zeroKL}
\label{stm:zero_kl}
    Let $\rposterior$ be the reparametrised posterior distribution defined by the density 
    $\dens{p} (\priorParam \given \dataset) 
    $.
    Assume the Gaussian prior and likelihood with $\lstd > 0$, and that $\smash{\enngp \overset{\Prob}{\to} \nngp}$ 
    and $\E [ \enngp ] \to \nngp$
    under the prior as $\minWidth \to \infty$.
    
    Then $\KL\pp{\gauss(0, I_{\dimParam}) \, \| \, \rposterior} \to 0$ as 
    $\minWidth \to \infty$.
\end{restatable}
The assumptions ($\smash{\enngp \overset{\Prob}{\to} \nngp}$, $\E [ \enngp ] \to \nngp$, as $\minWidth \to \infty$) hold for most common architectures,
including FCNs, CNNs, and attention networks \citep{matthews2018gaussian,alonso2018deep,hron2020infinite,yang2020tensor}.


The fact that KL divergence does not increase under measurable transformations \citep[e.g.,][corollary~5.2.2]{gray2011entropy}, combined with \Cref{stm:zero_kl}, 
implies that the KL divergence between the distribution of $\posteriorMap(\priorParam)$ with $\priorParam \sim \gauss(0, I_\dimParam)$ and the posterior $\posterior$
also converges to zero.
This provides a simple way of approximating wide BNN posteriors as illustrated by \Cref{fig:reparam_convergence}.
We use the same argument to establish convergence in \emph{function space} by showing measurability of the $\param \mapsto \fn_{\param}$ mapping 
w.r.t.\ a suitable $\sigma$-algebra.
\begin{restatable}{proposition}{klFunctions}
\label{stm:kl_functions}
    Let $\phi \sim \gauss( 0 , I_{\dimParam})$, $\param \sim \posterior$, and denote the functions they parametrise by $\fn_{\posteriorMap(\phi)}$ and $\fn_\param$.
    Assume all nonlinearities are continuous, and each layer's outputs are jointly continuous in the layer parameters and inputs.
    
    Then $\KL ( \distr{P}_{\fn_{\posteriorMap(\priorParam)}} \, \| \, \distr{P}_{\fn_{\param} \given \dataset} ) \to 0$ as 
    $\minWidth \to \infty$, where the $\KL$ is defined w.r.t.\ the usual product $\sigma$-algebra on $\R{\inputSpace}$.
\end{restatable}
\citet{hron2020exact} showed that for continuous bounded likelihoods, $\distr{P}_{\fn_\param \given \dataset}$ converges (weakly) to the NNGP posterior whenever $\distr{P}_{\fn_\param}$ with $\param \sim \distr{P}_{\param}$ converges to the NNGP prior.
This implies that the $\distr{P}_{\fn_{\posteriorMap(\priorParam)}}$ from \Cref{stm:kl_functions} converges (weakly) to the NNGP posterior whenever $\distr{P}_{\fn_{\param} \given \dataset}$ does:
by Pinsker's inequality, we have convergence in total variation which implies convergence of expectations of all bounded measurable (incl.\ bounded continuous) functions.

Our \Cref{stm:zero_kl} may moreover be seen as an appealing answer to the issue of finding a useful notion of `convergence' in parameter space of an \emph{increasing} dimension from \citep[section~4]{hron2020exact}.
As the authors themselves point out, their approach of embedding the weights $\param$ in $\R{\natnum}$, and studying weak convergence w.r.t.\ its usual product $\sigma$-algebra, `tells us little about behaviour of \emph{finite} BNNs'.
This is most clearly visible in their proposition~2, which establishes asymptotic reversion of the parameter space posterior to the \emph{prior} (without any reparametrisation), which we know does not induce the correct function space limit.

We note that our \Cref{stm:zero_kl} does not contradict the \citeauthor{hron2020exact}'s result
exactly because their
definition of convergence essentially only captures the \emph{marginal} behaviour of finite weight subsets, whereas our $\KL$ divergence approach characterises the \emph{joint} behaviour of all the weights.
Indeed, our next proposition shows that reversion to the prior does not occur under our stronger notion of convergence.
\begin{restatable}{proposition}{klPrior}
\label{stm:kl_prior}
    Under the assumptions of \Cref{stm:zero_kl}
    \begin{align*}
        &\KL \pp{\, \gauss(0, I_{\dimParam}) \, \| \, \posterior \,}
        \\
        &\to \tfrac{1}{2} [
            \lstd^{-2} \trace(\nngp)
            + \| y \|_{\lstd^{-2} I_\nPoints - \nngp_{\lstd^2}^{-1}}^2
            - \log \det (\lstd^{-2}\nngp_{\lstd^2})
        ]
        \nonumber
        \, ,
    \end{align*}
    as $\minWidth \to \infty$.
    The limit is equal to the KL divergence between the NNGP prior and posterior in function space,
    defined w.r.t.\ the usual product $\sigma$-algebra on $\R{\inputSpace}$.
\end{restatable}

\begin{figure*}[tbp]
    \centering
  \begin{tikzpicture}
    \node[anchor=south west,inner sep=0] (image) at (0,0) {\includegraphics[keepaspectratio,width=0.48\textwidth]{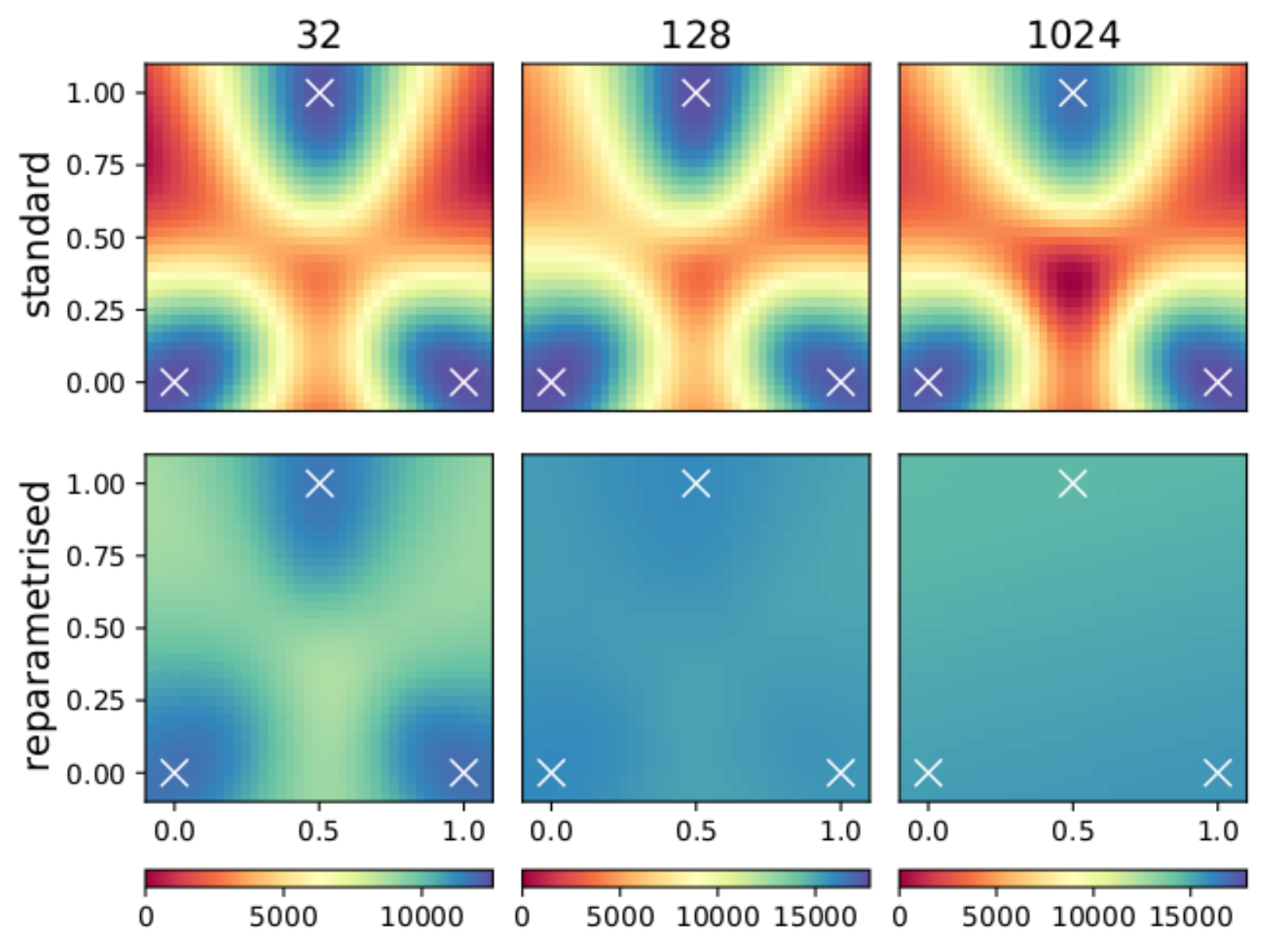}};
    \begin{scope}[x={(image.south east)},y={(image.north west)}]
    \node at (0.55, 1.025) {{\footnotesize \textbf{width}}};
    \end{scope}
  \end{tikzpicture}
  \hfill
  \begin{tikzpicture}
    \node[anchor=south west,inner sep=0] (image) at (0,0) {\includegraphics[keepaspectratio,width=0.48\textwidth]{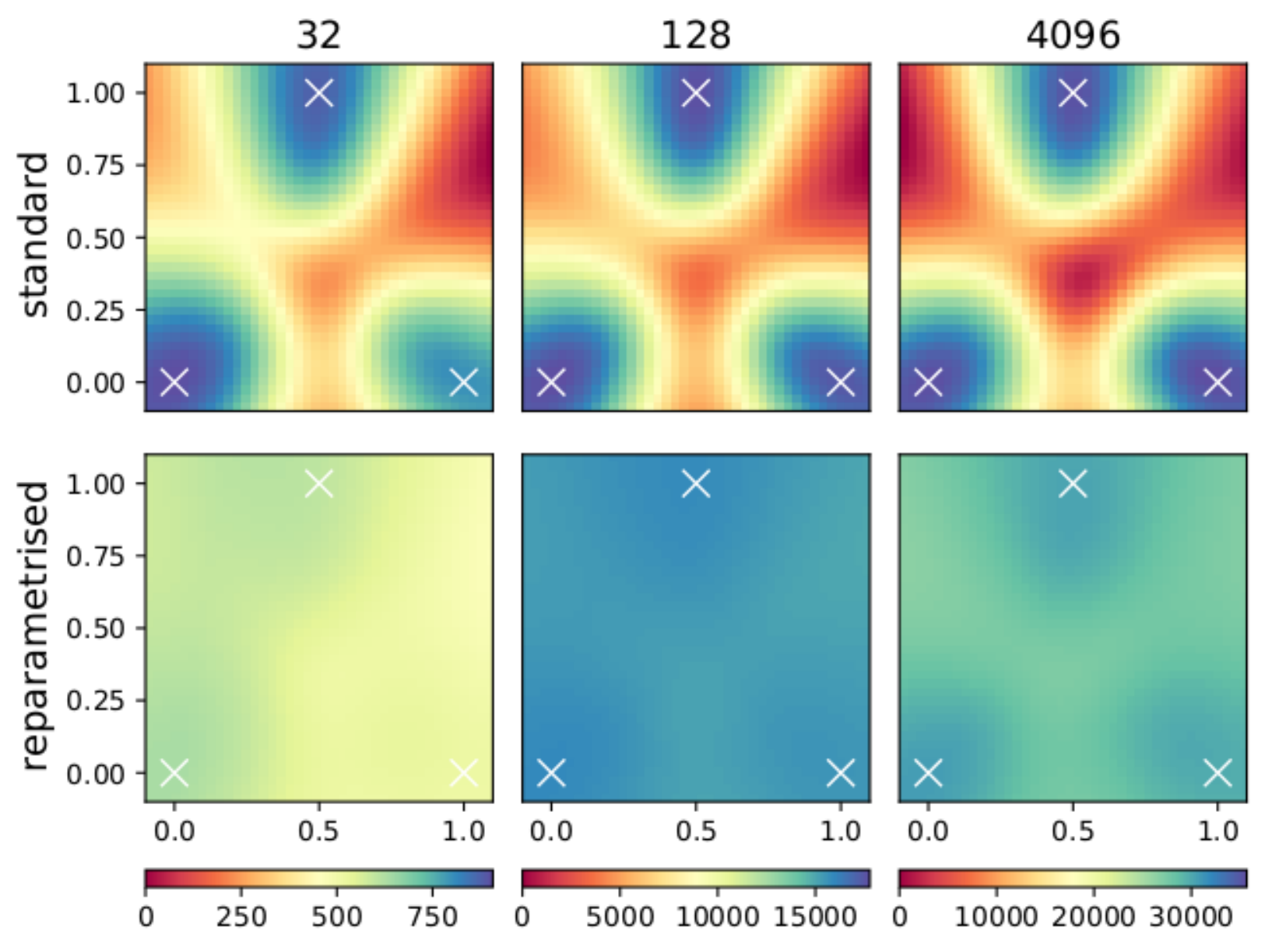}};
    \begin{scope}[x={(image.south east)},y={(image.north west)}]
    \node at (0.55, 1.025) {{\footnotesize \textbf{dataset size}}};
    \end{scope}
  \end{tikzpicture}
    \caption{
        \textbf{The posterior energy landscape is smoother after repriorisation.} 
        Plots show 2D slices of the log posterior on \texttt{cifar-10}
        in terms of the original parameters $\param$ (top row) and the reparametrised $\priorParam = T^{-1}\pp{\param}$ (bottom row), 
        for a 1-hidden layer FCN. 
        The 2D slices are obtained by spherical linear interpolation between three parameter values found by gradient descent training on CIFAR-10 after random initialisation; norms may differ, and are therefore interpolated linearly.
        \textbf{Left:} Columns represent varying hidden layer width. The number of observations is fixed at $\nPoints = 128$.
        The vanishing structure of the reparametrised posterior as width increases, making mixing between the normally separated modes easier.
        \textbf{Right:} Columns represent varying dataset size, width is fixed at $\dimParam^1 = 128$. As the number of datapoints becomes large compared to $\dimParam^1$, the reparametrised posterior gradually grows less smooth.
    }
    \label{fig:posterior_contour}
\end{figure*}

\section{Faster mixing via repriorisation}
\label{sect:sampling}

We now have a reparametrisation which \emph{provably} makes the reparametrised posterior close to the standard normal prior $\gauss(0, I_\dimParam)$.
Since sampling from the BNN posterior is notoriously difficult, the closeness to $\gauss(0, I_\dimParam)$ suggests sampling from $\rposterior$ may be significantly easier.

However closeness in KL divergence does not guarantee that the gradients of $\dens{p}(\priorParam \given \dataset)$ are as well behaved as those of the standard normal distribution,
which is crucial for gradient-guided methods like Langevin Monte Carlo (LMC).
In \Cref{sect:gradients}, we provide partial reassurance by proving that the $\ell^2$-norm of the differences between the two gradients vanishes in probability as the layers grow wide.
\Cref{sect:cholesky} then provides a simple-to-implement way of computing both the reparametrised $\param = \posteriorMap(\priorParam)$ and the corresponding density $\dens{p}(\priorParam \given \dataset)$ at a fraction of cost of the naive implementation.

\subsection{Convergence of density gradients}
\label{sect:gradients}

\looseness=-1
Our second major result proves that the \emph{posterior} mass of the region where $\nabla_{\priorParam} \log \dens{p}(\priorParam \given \dataset)$ significantly differs from the gradient of the $\gauss(0, I)$ density
vanishes in wide BNNs.
\begin{restatable}{proposition}{gradientConvergence}
\label{stm:gradient_convergence}
    Let $\Delta_\priorParam \coloneqq - \priorParam - \nabla_{\priorParam} \log \dens{p}(\priorParam \given \dataset)$ where $- \phi$ is the gradient of the log density of $\gauss(0, I)$.
    Assume the conditions in \Cref{stm:zero_kl}, and continuous nonlinearities.
    
    Then $\rposterior(\| \Delta_{\priorParam} \|_2 > \epsilon) \to 0$ as $\minWidth \to \infty$, $\forall \epsilon > 0$.
\end{restatable}
Two more technical assumptions are in \Cref{app:grad_convergence}.
Albeit useful, 
\Cref{stm:gradient_convergence} does not ensure
gradient-guided samplers like LMC
will not dawdle in
regions with ill-behaved gradients.
This behaviour did not occur in our experiments,
but the caveat is worth remembering.

\looseness=-1
To provide a rough quantitative intuition for the scale of the speed up, we show our reparametrisation enables $\nicefrac{\sqrt{\nPoints}}{\lstd}$-times higher LMC stepsize relative to no reparametrisation in \emph{linear} models, without any drop in the integrator accuracy (see \Cref{app:lmc_speedup_lin}).

\subsection{Practical implementation}\label{sect:cholesky}

To apply LMC, we need an efficient way of computing both $\param = \posteriorMap(\priorParam)$ and the corresponding gradient of $\log \dens{p}(\priorParam \given \dataset)$.
Since our largest experiments require millions of steps, the naive approach of computing each of $\mu$, $\Sigma^{1/2}$, and $\det (\Sigma^{1/2})$ separately---see \Cref{eq:determinant,eq:full_reparam}---
for the cost of \emph{three} $\mathcal{O}[(\rdimParam)^3]$ operations is worth improving upon.
We propose a three-times more efficient alternative.

The trick is to first compute the Cholesky decomposition 
$\chonngp^\top \chonngp = \lstd^2 I_{\rdimParam} + \embnngp^\top \embnngp$, and then reuse $\chonngp$ in computation of all three terms.
This can be done by observing $\Sigma = \lstd^2 \chonngp^{-1} (\chonngp^{-1})^\top$, which implies $\Var[\sigma \chonngp^{-1} \varepsilon ] = \Sigma$ for $\varepsilon \sim \gauss (0, I)$.
We can thus use the \emph{alternative} reparametrisation
\begin{align*}
    \rparam
    =
    \chonngp^{-1} 
    [
        (\chonngp^\top)^{-1}
        \embnngp^\top
        y
        +
        \lstd
        \rpriorParam
    ]
    \, ,
\end{align*}
which uses the fact $\mu = (\lstd^2 I + \embnngp^\top \embnngp)^{-1} \embnngp^\top y$ together with the above definition of $\chonngp$.
The determinant can then be computed essentially for free since 
\begin{align*}
    \det (\partial_\priorParam \param) 
    \overset{\text{(i)}}{=}
    \det (\lstd \chonngp^{-1})
    \overset{\text{(ii)}}{=}
    \frac{\lstd^{\rdimParam}}{\det(\chonngp)}
    \overset{\text{(iii)}}{=}
    \frac{\lstd^{\rdimParam}}{\prod_i U_{ii}}
    \, ,
\end{align*}
where (i) is as in \Cref{eq:determinant}, (ii) uses standard determinant identities, and (iii) exploits that $\chonngp$ is triangular.

Because $U$ is upper-triangular, $U^{-1} v$ (resp.\ $(U^\top)^{-1} v$) can be efficiently computed by back (resp.\ forward) substitution for any $v$ \citep{voevodov}.
The sequential application of forward and backward substitution is known as the Cholesky solver algorithm \citep{cholesky}.
Since we would need to apply the Cholesky solver to compute $\mu = (\lstd^2 I + \embnngp^\top \embnngp)^{-1} \embnngp^\top y$ anyway,
we obtain
both the
reparametrisation
the density 
at essentially the same cost.

\Cref{fig:speed} demonstrates our approach introduces little wall-time clock overhead on modern accelerators relative to no reparametrisation when combined with LMC.
Since the per-step computational overhead is never larger than 1.5x in \Cref{fig:speed} and we observe $\geq$10x faster mixing in \Cref{sect:experiments}, the cost is more than offset by the sampling speed up.

\looseness=-1
The above (`feature-space') reparametrisation 
scales as $\mathcal{O}[(\rdimParam)^3]$ in computation and $\mathcal{O}[(\rdimParam)^2]$ in memory.
This is typically much better than exact (NN)GP inference which scales as $\mathcal{O}(\nPoints^3)$ in computation and $\mathcal{O}(\nPoints^2)$ in memory.
To enable future study of very wide BNNs ($\rdimParam \gg \nPoints$), we provide an alternative `data-space' formulation with $\mathcal{O}(\nPoints^3)$ computation and $\mathcal{O}(\nPoints^2)$ memory scaling in \Cref{app:data_space_reparametrisation}.

\subsubsection*{Interpolating between parametrisations}

We can swap the likelihood variance $\lstd^2$ for a general regulariser $\kernelReg > 0$ in the definition of $\chonngp^\top \chonngp \coloneqq \kernelReg I_{\rdimParam} + \embnngp^\top \embnngp$.
Clearly, $\posteriorMap \to \mathrm{Id}$ (identity mapping) as $\kernelReg \to \infty$.
This allows interpolation between our proposed reparametrisation at $\kernelReg = \lstd^2$, and the standard parametrisation at $\kernelReg = \infty$, which may be useful when far from the NNGP regime.
Proper tuning of the hyperparameter $\lambda$ thus ensures the algorithm never performs worse than under the standard parametrisation.

\section{Experiments}\label{sect:experiments}

Armed with theoretical understanding and an efficient implementation, we now 
demonstrate that our reparametrisation can dramatically improve BNN posterior sampling, as quantified by per-step effective sample size \citep[ESS;][]{ripley1989}, and the $\hat{R}$ statistics \citep{gelman1992inference}, over a range of architectural choices and dataset sizes.

\subsection{Setup}\label{sect:experimental_setup}

\subsubsection{\texorpdfstring{$\hat{R}$}{R-hat} diagnostic}
\label{sect:rhat}

$\hat{R}$ is a standard tool for testing non-convergence of MCMC samplers.
It takes $M$ collections of samples $\{ z_{mi} \}_{i=1}^S \subset \R{}$ produced by \emph{independent} chains, and computes 
\begin{align}\label{eq:rhat_definition}
    \hat{R}^2
    =
    \frac{
        \hat{\E} \bigl[ \hat{\Var}(z_{mi} \given m) \bigr]
        + 
        \hat{\Var} \bigl[ \hat{\E}(z_{mi} \given m) \bigr]
    }{
        \hat{\E} \bigl[ \hat{\Var}(z_{mi} \given m) \bigr]
    }
    \, ,
\end{align}
where $\hat{\E}$ and $\hat{\Var}$ compute expectation and variance w.r.t.\ the empirical distribution which assigns a $\tfrac{1}{MS}$ weight to each $z_{mi}$ (conditioning on $m$ yields within-chain statistics).
The numerator is just $\hat{\Var}(z_{mi})$ by the law of total variance, where $\hat{\Var} \bigl[ \hat{\E}(z_{mi} \given m) \bigr]$ vanishes when all chains sample from the same distribution.
Values significantly larger than one thus indicate non-convergence.
We follow \citet{izmailov2021bayesian} in reporting $\hat{R}^2$ (denoted by $\hat{R}$ in their paper), instead of the square root of the above quantity.
For multi-dimensional random variables, we estimate $\hat{R}^2$ for each dimension, and report the resulting distribution over dimensions.

\begin{figure}[tbp]
    \centering
    \includegraphics[keepaspectratio,width=\linewidth]{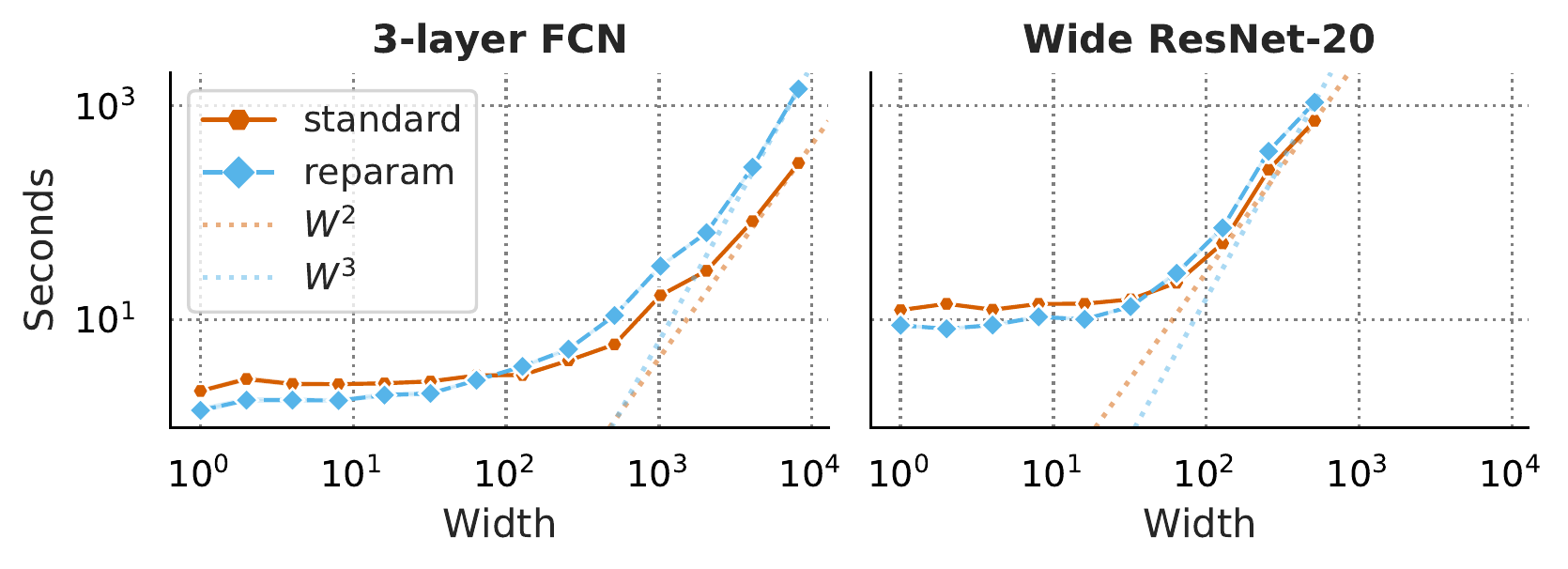}
    \caption{\looseness=-1
    \textbf{Sampling with repriorisation is comparable to the baseline in wall-clock time.} While our reparametrisation cost is cubic in top layer width $\rdimParam$ (\cref{sect:cholesky}),
    it is notable only for very large FCNs (\textbf{left}), and even larger CNNs (\textbf{right}) where the quadratic cost of the forward pass dominates due to a large multiplicative constant.
    See \Cref{app:data_space_reparametrisation} for an alternative formulation which scales only quadratically with $\dimParam^\nLayers$ but cubically with $\nPoints$.
    }
    \label{fig:speed}\vspace{-1em}
\end{figure}

\subsubsection{Effective sample size}\label{sect:ess}

ESS is a measure of sampling speed.
It estimates lag $k$ autocorrelations $\hat{\rho}_k \coloneqq \hat{\Covar} (z_{mi}, z_{m (i+k)}) / \, \hat{\Var}(z_{mi})$ for $k = 1, \ldots, S - 1$ ($\hat{\Covar}$ is the empirical covariance), and computes
\begin{align}\label{eq:ess_definition}
    \widehat{\text{ESS}} 
    \coloneqq
    \frac{S}{
        1 + 
        2 
        \sum_{k=1}^{S - 1}
            (1 - \tfrac{k}{S}) \hat{\rho}_k
    }
    \, ,
\end{align}
The per-step ESS is then $\nicefrac{\widehat{\text{ESS}}}{S}$.
Since $\hat{\rho}_k$ estimates for high $k$ are based on only $M (S - k)$ samples, we denoise by stopping the sum in \Cref{eq:ess_definition} at the first $\hat{\rho}_k < 0$, which is the default in Tensorflow Probability's \href{https://www.tensorflow.org/probability/api_docs/python/tfp/substrates/jax/mcmc/effective_sample_size}{\texttt{effective\_sample\_size}} \citep{dillon2017tensorflow}.

\Cref{eq:ess_definition} is based on the $\Var(\bar{z}_S) = \nicefrac{\Var(z_1)}{S}$ identity for (population) variance of the empirical mean $\bar{z}_S$ of $S$ \emph{i.i.d.}\ variables $z_i$
Since MCMC chains produce dependent samples, $\smash{\widehat{\text{ESS}}}$ estimates the
number which would satisfy the above equality when substituted for $S$ under the assumption of stationarity.
Because we need to estimate ESS in the very high-dimensional parameter space,
we randomly choose \emph{100 one-dimensional subspaces}, project the samples, estimate ESS in each of them,
and report their distribution.

\begin{figure*}[htbp]
    \centering
  \begin{tikzpicture}
    \node[anchor=south west,inner sep=0] (image) at (0,0) {
    \includegraphics[keepaspectratio,width=0.45\textwidth]{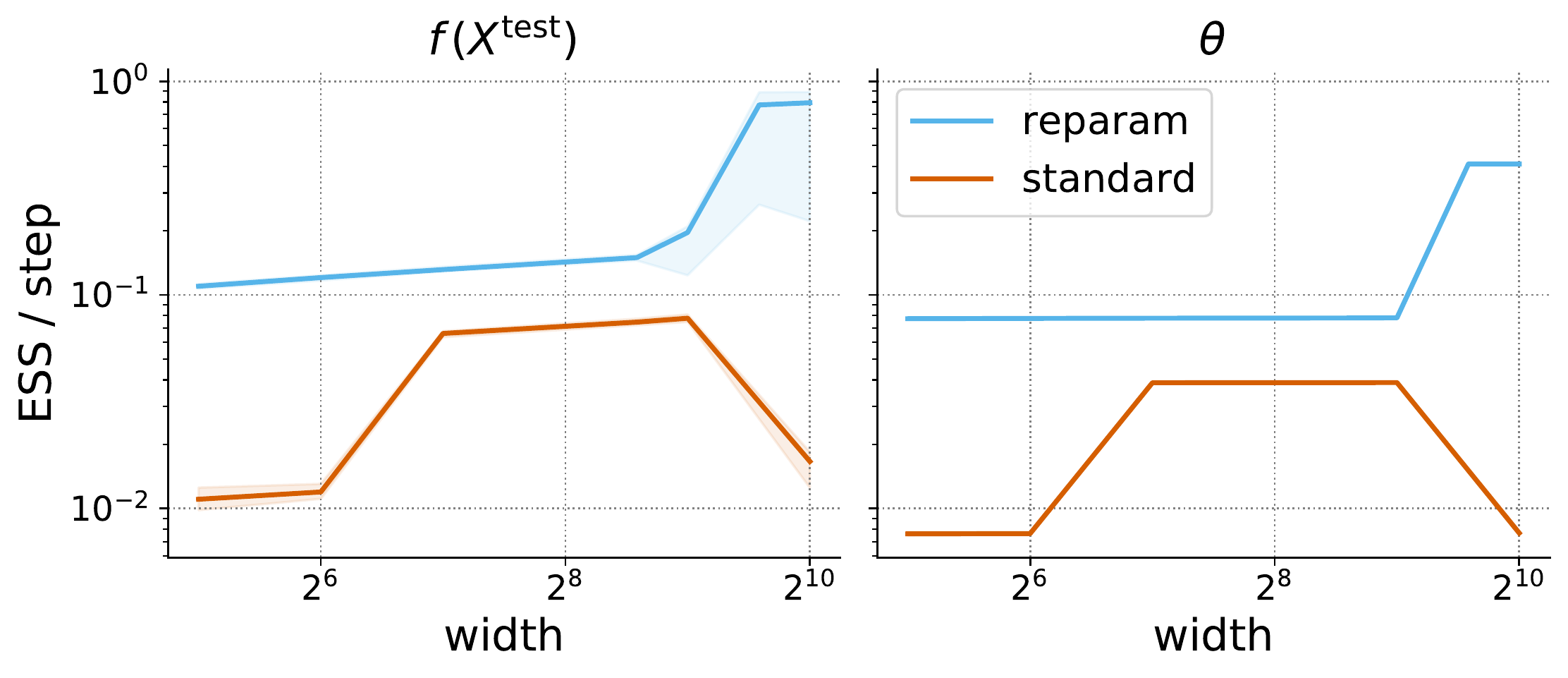}
    };
    \begin{scope}[x={(image.south east)},y={(image.north west)}]
    \node at (0.55, 1.075) {{\footnotesize \textbf{width}}};
    \end{scope}
  \end{tikzpicture}
  \hfill
  \begin{tikzpicture}
    \node[anchor=south west,inner sep=0] (image) at (0,0) {
    \includegraphics[keepaspectratio,width=0.45\textwidth]{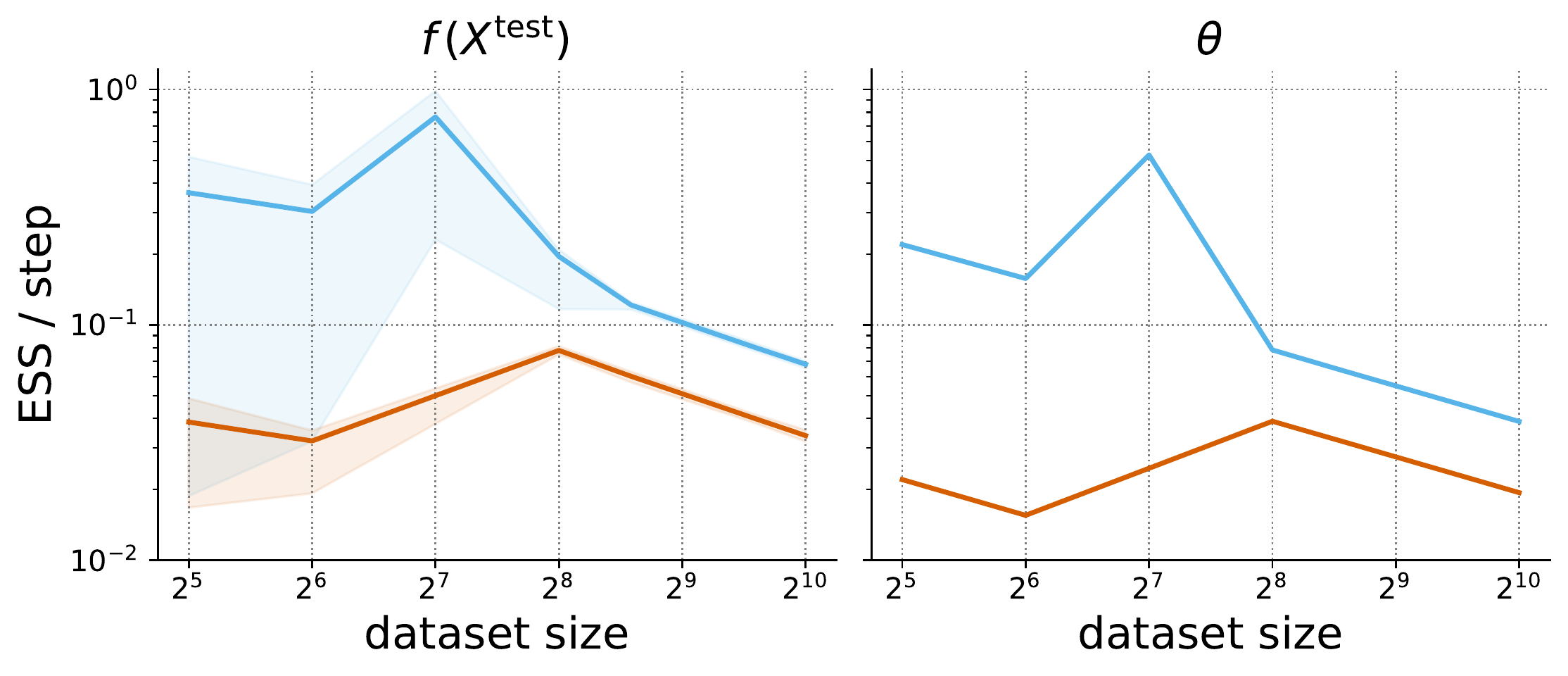}
    }; 
    \begin{scope}[x={(image.south east)},y={(image.north west)}]
    \node at (0.55, 1.075) {{\footnotesize \textbf{dataset size}}};
    \end{scope}
  \end{tikzpicture}
    \vspace*{-0.75em}
    \caption{\textbf{Repriorisation makes mixing speed faster, and the performance benefits increase as the width to dataset size ratio increases.} Per-step ESS as a function of layer width (left) and dataset size (right)
    for a 3-hidden layer FCN.
    The line indicates the mean, and the bands the \emph{minimum} and \emph{maximum}, ESS over the $100$ projection subspaces (see \Cref{sect:ess}).
    In all cases, reparametrisation achieves significantly better mixing speed (note the log-log scale of the axes).
    \texttt{3M} samples collected for each configuration.
    Hyperparameters were tuned for each configuration separately (see \Cref{sect:width_dataset_experiments}).
    \textbf{Left:} Dataset size fixed at $\nPoints = 256$.
    The benefit of reparametrisation increases with width, especially so when width becomes larger than dataset size where reparametrisation yields 50x higher ESS.
    \textbf{Right:} Width fixed at $512$.
    Note that reparameterisation leads to a higher ESS for all configurations.}
    \label{fig:ess_width_dataset}
\end{figure*}

\subsubsection{Data and hyperparameters}\label{sect:data_hypers}

\begin{figure*}[htb]
    \centering
  \begin{tikzpicture}
    \node[anchor=south west,inner sep=0] (image) at (0,0) {
    \includegraphics[keepaspectratio,width=0.48\textwidth]{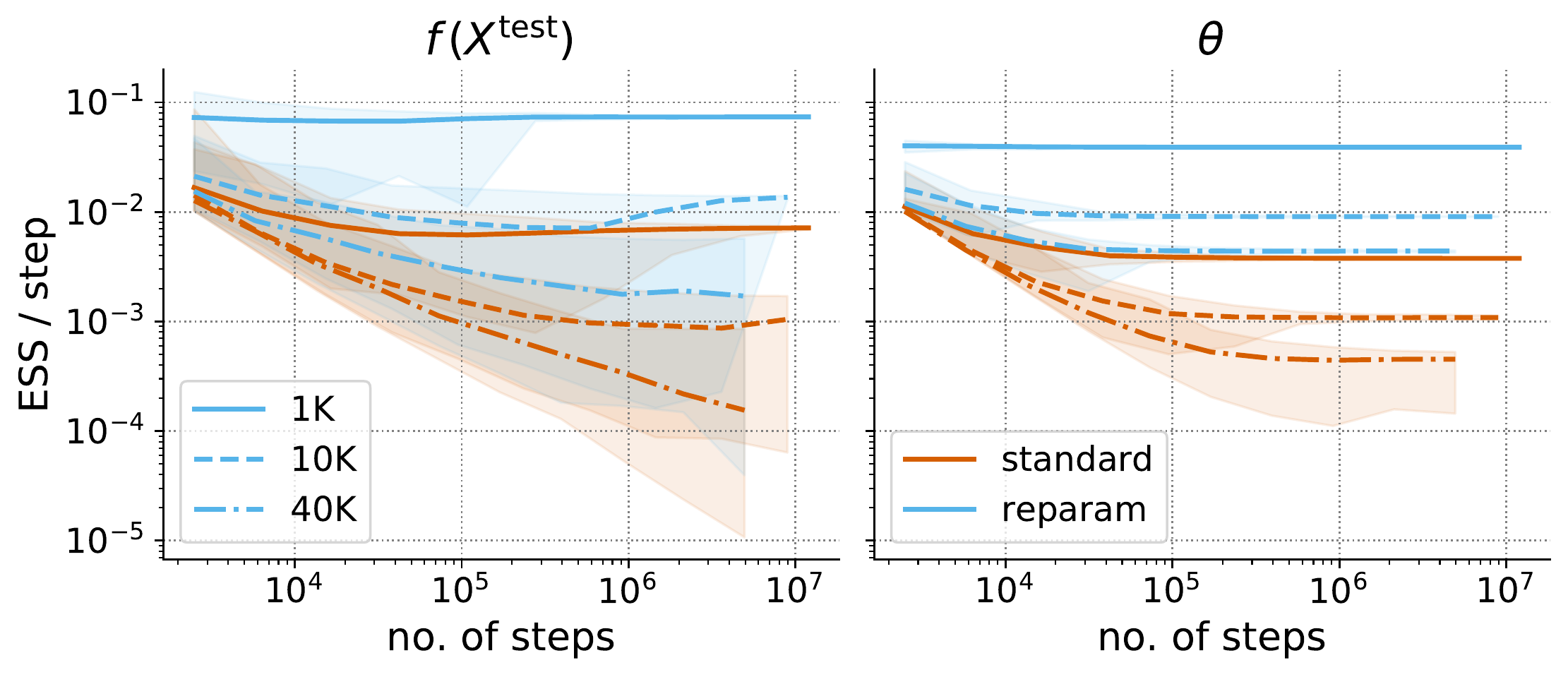}
    };
    \begin{scope}[x={(image.south east)},y={(image.north west)}]
    \node at (0.55, 1.075) {{\footnotesize \textbf{3-hidden layer FCN}}};
    \end{scope}
  \end{tikzpicture}
  \hfill
  \begin{tikzpicture}
    \node[anchor=south west,inner sep=0] (image) at (0,0) {
    \includegraphics[keepaspectratio,width=0.48\textwidth]{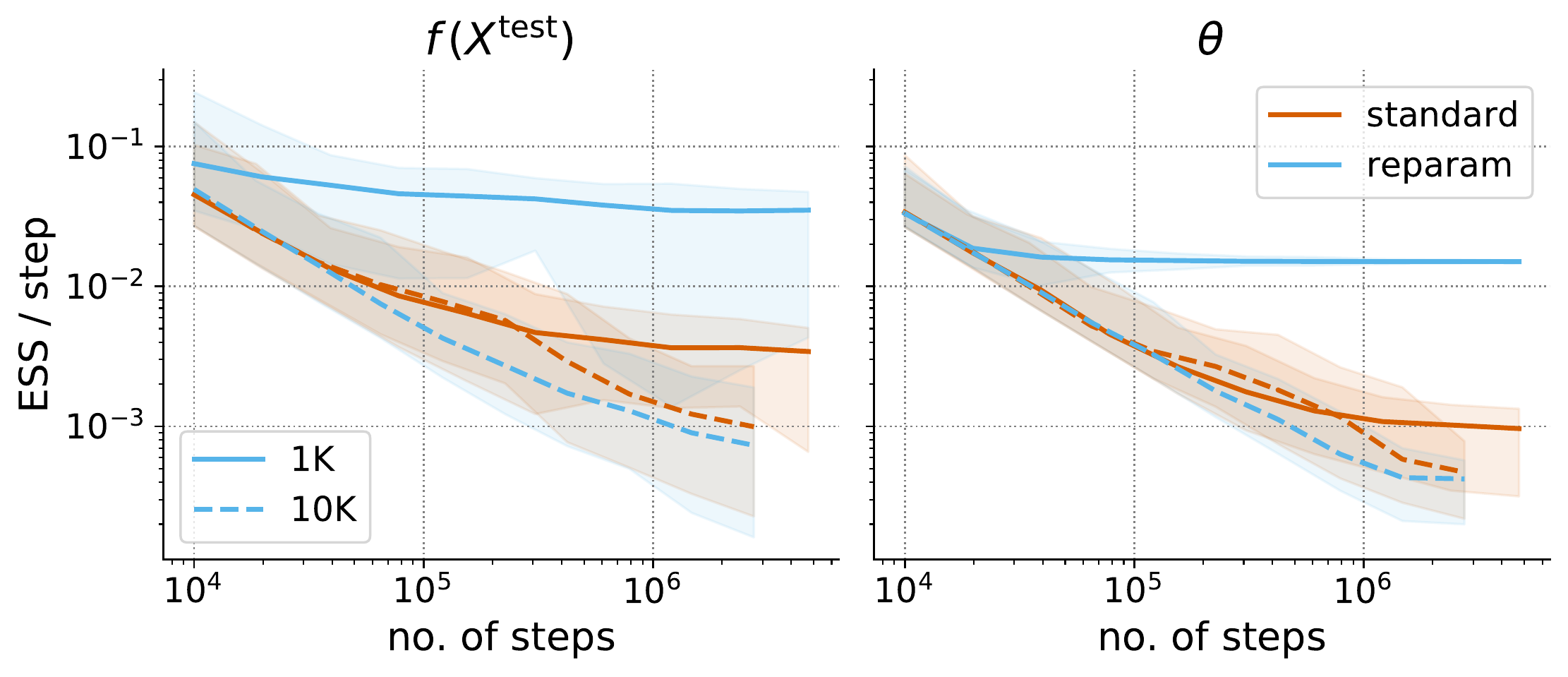}
    }; 
    \begin{scope}[x={(image.south east)},y={(image.north west)}]
    \node at (0.55, 1.075) {{\footnotesize \textbf{Wide ResNet-20}}};
    \end{scope}
  \end{tikzpicture}
    \caption{\textbf{Repriorisation results in faster mixing across architectures.} 
    Per-step ESS estimates are plotted as a function of the number of sampler steps for three random subsets of \texttt{cifar-10} of size $\nPoints = 1024, 10240, 40960$.
    The line indicates the mean, and the bands the \emph{minimum} and \emph{maximum}, ESS over the $100$ projection subspaces (see \Cref{sect:ess}).
    The initial downward trend of the curves is due to underestimation of long range autocorrelations at earlier steps where sample size is too small. \textbf{Left:} Results for a FCN with three width 1024 hidden layers. 
    The non-reparametrised BNNs mix around 10x slower for all three dataset sizes,
    and up to 200x slower if the worst mixing projections (bottom of shaded regions) are compared to each other for each case.
    \textbf{Right:} Results for a normaliser-free Wide ResNet-20 with 128 and 512 units (channels) in the narrowest and widest layers (see \Cref{sect:architecture_experiments}). 
    For $\nPoints =$ \texttt{1K}, reparameterisation performs more than 10x better. 
    For $\nPoints =$ \texttt{10K}, both chains have very similar ESS / step values. 
    This could be because the dataset size is now two orders of magnitude larger than the width, a regime in which our theory does not apply. 
    It could also be because, even after 3 million sampling steps, the chains are so far from equilibrium and the ESS estimates are dominated by the initial transient effects.
    }
    \label{fig:ess_three_subsets}
\end{figure*}

\looseness=-1
The underdamped LMC sampler \citep{rossky1978} and \texttt{cifar-10} dataset \citep{krizhevsky2009learning} are used in all experiments.
We use regression on one-hot encoding of the labels shifted by $-\tfrac{1}{10}$ so that the label mean of each example is zero, matching the BNN prior (extension to multi-dimensional outputs described in \Cref{app:multi_dim}).
Gaussian likelihood with $\lstd = 0.1$ is used throughout, which we selected to ensure that (a)~positive and negative labels are separated by at least $4\lstd$, but (b)~$\lstd$ is not too small to give ourselves an unfair advantage based on the linear case intuition (\Cref{app:lmc_speedup_lin}), or cause numerical issues.
Sample thinning is applied in all experiments, 
and ESS and $\hat{R}$ are always computed based on the thinned sample.
However, \Cref{fig:ess_three_subsets,fig:rhat_three_subsets} visualise ESS evolution as a function of the number of LMC steps which is equal the \emph{unthinned} sample size.

\looseness=-1
As common, we omit the Metropolis-Hastings correction of LMC \citep{Neal93probabilisticinference}, and instead tune hyperparameters so that average acceptance probability after burn-in stays above 98\%.
We observed significant impact of numerical errors on both the proposal and the calculation of the acceptance probability;
we use \texttt{float32} which offers significant speed advantages over \texttt{float64}, but can still suffer 2-5 percentage point changes in acceptance probability compared to \texttt{float64} computation with the same candidate sample.

Our experiments are implemented in JAX \citep{jax2018github}, and rely on the Neural Tangents library \citep{neuraltangents2020} for both implementation of the NN model logic, and evaluation of the NNGP predictions (e.g., in \Cref{fig:reparam_convergence}).\footnote{Code: \href{https://github.com/google/wide_bnn_sampling}{github.com/google/wide\_bnn\_sampling}.}

\subsection{Results}

In Figure \ref{fig:posterior_contour} we visualize 2D slices through the weight posterior, for the original and reparameterized network for a 1-hidden layer FCN. The log posterior is far smoother after reparametrization, and its relative smoothness improves as network width is increased relative to dataset size. As we quantify in the following sections, this enables dramatic improvements in sampling efficiency.

\subsubsection{Dependence on width and dataset size}
\label{sect:width_dataset_experiments}

In \Cref{fig:ess_width_dataset}, we explore the dependence of mixing speed of the LMC sampler on network width and dataset size as measured by the per-step ESS.
We use a single chain run of \texttt{3M} steps with a \texttt{20K} burn-in per each width/training set size.
The \emph{results are computed on a sample thinned by a factor of 25}.
We use a FCN with three equal width hidden layers and GELU nonlinearities \citep{hendrycks2016gaussian}, and the weight scaling described in \Cref{sect:notation_assumptions}.
The weight and bias scaling factors are set to $\wstd^2 = 2$, $\bstd^2 = 0.01$ everywhere except in the readout layer where $\wstd^2 =  1$ to achieve approximately unit average variance of the network predictions under the prior (at initialisation).
For the width experiments (\Cref{fig:ess_width_dataset}, left), the dataset size was fixed at $\nPoints = 256$, and the layer width varied over $\{ 32, 64, 128, 384, 512, 768, 1024 \}$.

For each width and both parametrisations, we \emph{separately} tuned the LMC stepsize and damping factor to maximise the ESS of $\param$ while satisfying the $>98\%$ mean acceptance probability criterion.
For the reparametrised sampler, we also tried to tune the regularisation parameter $\kernelReg$ (see \Cref{sect:cholesky}), but found that the value dictated by our wide BNN theory ($\kernelReg = \lstd^2$) worked best even for the smallest width models.
The distribution of ESS over the random projections of $\param$ and $\fn(X^\text{test})$ on \texttt{5K} test points is shown, with its mean, minimum, and maximum values.

Our reparametrisation achieves significantly faster mixing in all conditions (note the log-log scale).
The improvement becomes larger as the BNN grows wider (left), achieving near-perfect values for the widest BNNs ($10^0=1$ is maximum) where the width ($>\! 2^{8}$) is larger than the dataset ($256 = 2^8$), with a 50x speed up at width 1024.

For the dataset size plots (right), the performance is again best when the ratio of width to the number of observations is the smallest (here on the left).
The ESS of both $\fn(X^\text{test})$ and $\param$ decays with dataset size as expected, since the posterior becomes more complicated, and the width to dataset size ratio dips below one, i.e., we are leaving the NNGP regime. 
The reparametrisation however maintains higher ESS even outside of the conditions where our theory holds.

\subsubsection{Dependence on architecture}
\label{sect:architecture_experiments}

While our theory does not cover the cases where the ratio of dataset size to width is large, the consistent advantage of reparametrisation even in this regime is intriguing (\Cref{fig:ess_width_dataset}, right).
In this section, we investigate the dependence of this phenomenon on the architecture and dataset size.

For the architecture, we compare the FCN from \Cref{sect:width_dataset_experiments} with a normaliser-free Wide ResNet-20 with 128, 256, and 512 channels arranged from bottom to the top layer in the usual three groups of residual layer blocks \citep{he2016resnet,zagoruyko2016wideresnet}.
GELU is used for both.

As in \citep{izmailov2021bayesian}, we remove batch normalisation since it makes predictions depend on the mini-batch, which interferes with Bayesian interpretation.
We however use a different alternative in its place, namely the proposal of \citet{shao2020normalisation}, which replaces each residual sum $\fn_k^\text{skip}(x) + \fn_k^\text{resid}(x)$ by $\sqrt{\nicefrac{k - 1 + c}{k + c}} \, \fn_k^\text{skip}(x) + \sqrt{\nicefrac{1}{k + c}} \, \fn_k^\text{resid}(x)$ where $k$ is the index of the skip connection, and $c$ is a hyperparameter we set to $1$.
Since \citep{izmailov2021bayesian} is the closest comparison for this section, we note the authors use a categorical instead of a Gaussian likelihood\footnote{Our reparameterization also enables accelerated sampling for categorical likelihood -- see Appendix \ref{app:non-gauss_likelihood}.}, and an i.i.d.\ $\gauss(0, \alpha)$ prior for all parameters, but \emph{without} scaling the weights by $\nicefrac{\wstd^\layer}{\sqrt{\dimParam^\layer}}$ as we do (see \Cref{sect:notation_assumptions}).

For the scaling factors, FCN is as in \Cref{sect:width_dataset_experiments}, whereas the ResNet uses $\wstd^2 = 2.2$ and $\bstd^2 = 0$, except in the last layer where $\wstd^2 = 1$ and $\bstd^2 = 0.01$ is again employed to ensure reasonable scale of the network predictions under the prior. 
As a sanity check, using this prior as initialiser, and optimising with Adam, this normaliser-free ResNet achieves a decent $95.6\%$ validation accuracy on full \texttt{cifar-10}.
We tuned the LMC stepsize and damping factor separately for every configuration to maximise ESS of $\param$ while satisfying $> 98$\% mean acceptance rate after burn-in.
For the reparametrised chain, the regulariser $\kernelReg$ was $\lstd^2$ and $\nPoints \lstd^2$ respectively for the FCN and the ResNet, with $\lstd^2 = 0.01$ the output variance.
The \emph{sample thinning factor is 100}.

\begin{figure}[tbp]
    \centering
    \includegraphics[keepaspectratio,width=0.925\linewidth]{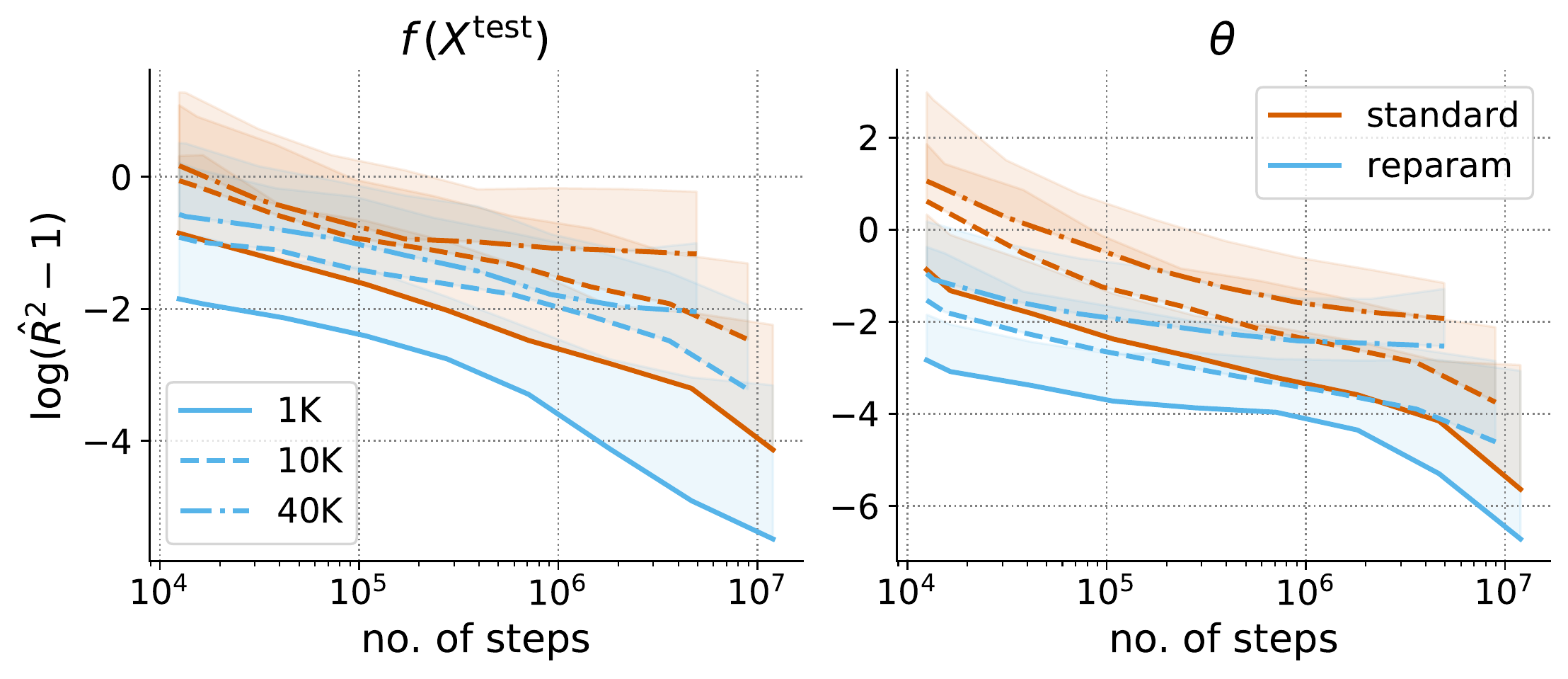}
    \caption{\textbf{Convergence of $\hat{R}$ with varying dataset size.} Complements the left side of \Cref{fig:ess_three_subsets}. 
    Line represents mean, band the maximum value.
    Estimates based on three independent chains. Note the log-log scale, and 
    that $\hat{R}^2$ is plotted relative to 
    its unit lower bound. 
    High $\hat{R}^2$ indicate non-convergence so lower is better.}
    \label{fig:rhat_three_subsets}\vspace{-1em}
\end{figure}

As can be seen in \Cref{fig:ess_three_subsets}, the mixing speed advantage of reparametrisation persists far from the NNGP regime in the case of the FCN (left), where even a factor of forty in dataset to width ratio did not erase the surprisingly consistent 10x higher (average) ESS.
This observation is supported by \Cref{fig:rhat_three_subsets}, in which $\hat{R}^2$ is already near one (optimum) when the standard parametrisation still exhibits $\mathcal{O}(10^2)$ values.
 
For the Wide ResNet (right), we observed a similar $\geq$10x benefit from reparameterisation at \texttt{1K} datapoints. 
For the $n=$\texttt{10K} case, no benefit was observed. 
This may be because both chains are far from equilibrium even after \texttt{3M} steps, and the per-step ESS measurement reflects initial transients. 
It may also be because the dataset size (10,240) was much larger than smallest channel count (128), exceeding by two orders of magnitude the scale where the model is expected to approximate the NNGP limit.

\section{Other related work}


For ensembles of infinitely-wide NNs trained with gradient descent, \citet{shwartz2020information} find the KL divergence between the posterior and the prior to diverge linearly with training time, an interesting departure from the finite value we find for Bayesian inference in \Cref{stm:kl_prior}. The information content of infinitely-wide NNs was analysed by \citet{bernstein2021computing} using the NNGP posterior, leading to non-vacuous PAC-Bayes generalisation bounds, adding to an active line of work focusing on generalisation bounds for overparametrised models \citep[see, e.g.,][]{dziugaite2017computing,valle2018deep,vakili2021uniform}.

If some hidden layers are narrow and remain finite, the resulting model is a type of Deep Gaussian process~\cite{damianou2013deep,agrawal2020wide} or Deep kernel process~\cite{aitchison2021deep}. \citet{aitchison2020bigger} argues that such models may exhibit improved performance relative to uniformly wide BNNs owing to a form of representation learning. While we do not investigate such questions in this work, the favourable computational properties of our LMC algorithm could facilitate such analyses in the future. 

Moving away from the infinite width limit, \citet{yaida2020non} and \citet{roberts2021principles} show how to compute the Bayesian prior and posterior systematically in powers of $1/\text{width}$, with the leading-order term given by the NNGP, though the subleading corrections pose computational challenges. For some architectures, a Gaussian prior in parameter space leads to closed-form expressions for the prior in function space, even at finite width \cite{zavatone2021exact}. Less is known about the posterior for finite BNNs, since computational considerations often require approximate inference, whose quality in real-world problems has 
recently been investigated in \citep{izmailov2021bayesian}. 

Our work is partially inspired by a thread of papers which accelerate MCMC sampling by applying an invertible transformation to a distribution's variables, and then running sampling chains in the transformed space 
\citep[e.g.,][]{el2012bayesian,marzouk2014transport,parno2015transport,marzouk2016introduction,titsias2017learning,hoffman2019neutra}.

\section{Conclusion}

We introduced \emph{repriorisation}, a reparameterisation that (1)~provides a rigorous characterisation of wide BNN parameter space, and (2)~enables a more efficient BNN sampling algorithm which mixes \emph{faster} as the network is made wider. 


Our theory shows
BNN posteriors exhibit non-negligible interactions between parameters in different layers,
even at large width.
In contrast, many popular BNN posterior approximations (e.g., mean-field methods) assume independence between parameters in different layers.
Beyond MCMC,
algorithms which incorporate between-layer interactions---from Laplace \citep{tierney1986accurate} to more recent ones like \citep{ober2021global}---are therefore better positioned to capture the true BNN posterior behaviour.
In fairness, approximation fidelity and downstream performance are not the same thing though, as clearly demonstrated by the success of non-Bayesian approaches.

\looseness=-1
Our sampling results parallel the interplay between optimisation and model size in deterministic NNs. 
In particular, first-order methods used to be considered ill-suited for the non-convex loss landscapes of NNs, yet experimental and later theoretical results showed they can be highly effective, especially for large NNs \citep[e.g.,][]{allen2019convergence,du2019gradient}.
Similarly, MCMC theory tells us that sampling is often harder in high dimensions, yet our results show that a simple reparametrisation exploiting the particular structure of wide BNN posteriors makes sampling much easier.

While the gap between deterministic and Bayesian NNs remains considerable, we hope our work enables further progress for practical large-scale BNNs.





\bibliography{paper}
\bibliographystyle{icml2022}

\newpage
\appendix
\onecolumn


\section{Proofs}\label{app:proofs}

Here and in the main body of the paper, all statements of convergence in probability or almost sure (a.s.) convergence with $\minWidth \to \infty$ necessitate construction of an underlying probability space where the random variables \emph{for network of all widths} live.
Since verifying that the assumptions we use are satisfied for a majority of NN architectures can be obtained by referring to the Tensor Program work of \citeauthor{yang2020tensor}, we adopt his probability space construction \citep[e.g.,][appendix L]{yang2020tensor}.

Throughout we use the following definitions: $\lesssim$ is $\leq$ up to a universal positive (i.e., $a \lesssim b$ means $\exists c > 0$ s.t.\ $a \leq c \, b$), $\sdimParam \coloneqq \dimParam - \rdimParam = \dim (\spriorParam)$, $\langle r, s \rangle = r^\top s$ is the dot product,
$\regnngp[\lstd^2] \coloneqq \lstd^2 I_\nPoints + \enngp$,
and $\nngp_{\lstd^2} \coloneqq \lstd^2 I_\nPoints + \nngp$.

\subsection{\texorpdfstring{\Cref{stm:zero_kl}}{Zero parameter-space KL}}

\zeroKL*

\begin{proof}[Proof of \Cref{stm:zero_kl}]
    By the conditional KL divergence identity
    \begin{align*}
        \KL ( \gauss(0, I_{\dimParam}) \, \| \, \rposterior )
        =
        \KL ( \gauss(0, I_{\sdimParam}) \, \| \, \distr{P}_{\spriorParam \given \dataset} )
        + 
        \E_{\spriorParam \sim \gauss(0, I_{\sdimParam})} [
                \KL (\gauss(0, I_{\rdimParam}) \, \| \, \distr{P}_{\rpriorParam \given \spriorParam, \dataset})
        ]
        \, ,
    \end{align*}
    where the latter term is zero by the argument right after \Cref{eq:full_reparam}.
    Recalling $\spriorParam = \sparam$ and $\dens{p}(\sparam) \sim \gauss (0, I_{\sdimParam})$
    \begin{align*}
        &\KL ( \gauss(0, I_{\sdimParam}) \, \| \, \distr{P}_{\spriorParam \given \dataset} )
        =
        \log Z 
        +
        \tfrac{1}{2}
        \E_{\spriorParam \sim \gauss (0, I)} \bigl[
            \| y \|_{\regnngp[\lstd^2]^{-1}}^2
            +
            \log \det (\regnngp[\lstd^2])
        \bigr]
        \, ,
        \, ,
    \end{align*}
    where we substituted \Cref{eq:remainder_posterior} renormalised by $
    Z 
    \coloneqq 
    \E_{\spriorParam \sim \gauss (0, I_{\sdimParam})} 
        \exp \{
            - \tfrac{1}{2} [
                \| y \|_{\regnngp[\lstd^2]^{-1}}^2
                +
                \log \det (\regnngp[\lstd^2])
            ]
        \}
    $.
    Note that
    $$
        \enngp
        \mapsto
        \exp \{
            - \tfrac{1}{2} [
                \| y \|_{\regnngp[\lstd^2]^{-1}}^2
                +
                \log \det (\regnngp[\lstd^2])
            ]
        \}
        \, ,
    $$
    is a bounded function of $\enngp$ by the positive semidefiniteness of $\embnngp \embnngp^\top$.
    It is also continuous in $\enngp$:
    for the determinant use, e.g., the Leibniz formula; the matrix inverse is continuous on the space of invertible matrices by continuity of the determinant and of $A \mapsto \mathrm{adj}\, A$ (sufficient here since $\regnngp[\lstd^2] = \lstd^2 I_\nPoints + \enngp$ where $\enngp$ is positive semidefinite).
    Since the measure w.r.t. which we integrate is the prior $\gauss (0, I_{\sdimParam})$, and 
    $\enngp \to \nngp$ in probability under the prior by assumption
    \begin{align*}
        \log Z \to 
        - \tfrac{1}{2} [
            \| y \|_{\nngp_{\lstd^2}^{-1}}^2
            +
            \log \det (\nngp_{\lstd^2})
        ]
        \quad
        \text{as}
        \quad
        \minWidth \to \infty
        \, ,
    \end{align*}
    by the definition of convergence in distribution (implied by convergence in probability), and the continuity of the $\log$.
    
    All that remains is thus to prove
    $$
        \E_{\spriorParam \sim \gauss (0, I)} \bigl[
            \| y \|_{\regnngp[\lstd^2]^{-1}}^2
            +
            \log \det (\regnngp[\lstd^2])
        \bigr]
        \to
            \| y \|_{\nngp_{\lstd^2}^{-1}}^2
            +
            \log \det (\nngp_{\lstd^2})
        \quad
        \text{as}
        \quad
        \minWidth \to \infty
        \, .
    $$
    Note that the integrand is continuous in $\enngp$ by the above argument.
    Using the convergence of $\enngp$ to $\nngp$ in probability under the prior, we thus see that the integrand converges in probability to the above constant by the continuous mapping theorem.
    
    Convergence of the expectation can then be ensured by establishing uniform integrability (UI), and invoking the Vitali's convergence theorem for finite measures (or theorem~3.5 in \citealp{billingsley1999convergence}).
    This is simple for the first term since 
    $$0 \leq \| y \|_{\nngp_{\lstd^2}^{-1}}^2 \leq \lstd^{-2} \| y \|_2^2 \quad \text{a.s.,}$$
    which trivially implies UI.
    For the log determinant, note $0 \leq \log \det (\regnngp[\lstd^2]) \leq \trace(\regnngp[\lstd^2])$ (a.s.) by the arithmetic-geometric mean inequality.
    By \Cref{stm:ui_group_bound}, UI integrability can thus be obtained by establishing UI of $\{ \trace(\enngp) \}$ (indexed by width).
    Because $\trace(\enngp) \geq 0$ (a.s.) and $\E [\trace(\enngp) ] = \trace( \E [\enngp ] ) \to \trace( \E [\nngp ] ) = \E [ \trace(\nngp) ]$ under the prior by assumption, the desired UI follows by theorem~3.6 in \citep{billingsley1999convergence}.
    The UI of the sum then follows by the triangle inequality.
\end{proof}

\subsection{\texorpdfstring{\Cref{stm:kl_functions}}{Zero function-space KL}}

\klFunctions*

\begin{proof}[Proof of \Cref{stm:kl_functions}]
    As mentioned in the main text, KL divergence between the pushforwards of two distributions through the \emph{same} measurable mapping is never larger than that between the original distributions \citep[e.g.,][corollary~5.2.2]{gray2011entropy}.
    It is thus sufficient to show that for any given fixed network architecture, the mapping from the parameters into the function space $\param \in \R{\dimParam} \mapsto \fn_\param \in \R{\inputSpace}$ is measurable with respect to the usual Borel product $\sigma$-algebras on $\R{\dimParam}$ and $\R{\inputSpace}$.
    
    The product $\sigma$-algebra on $\R{\inputSpace}$ is by definition generated by the product topology.
    The product topology is generated by a base composed of sets $A = \bigcap_{i=1}^{k} \pi_{k}^{-1}(A_i)$, where each $\pi_k \colon \R{\inputSpace} \to \R{}$ is a coordinate projection associated with a point $x_k \in \inputSpace$, and each $A_k$ is an open set in the output space $\outputSpace \subseteq \R{}$.
    Because any finite intersection of open sets is open, it is sufficient to show that the mapping $\param \mapsto \fn_\param(x)$ is continuous for every $x \in \inputSpace$.
    Since the outputs of each layer are jointly continuous in the inputs and its own parameters by assumption, we can progress recursively from the input to the output, and conclude by observing that compositions of continuous functions are continuous.
\end{proof}

\subsection{\texorpdfstring{\Cref{stm:kl_prior}}{Non-zero constant KL from prior}}

\klPrior*

\begin{proof}[Proof of \Cref{stm:kl_prior}]
    By the conditional KL divergence identity
    \begin{align*}
        \KL ( \gauss(0, I_{\dimParam}) \, \| \, \posterior )
        =
        \KL ( \gauss(0, I_{\sdimParam}) \, \| \, \distr{P}_{\sparam \given \dataset} )
        + 
        \E_{\sparam \sim \gauss(0, I_{\sdimParam})} [
            \KL (\gauss(0, I_{\rdimParam}) \, \| \, \distr{P}_{\rparam \given \sparam, \dataset})
        ]
        \, .
    \end{align*}
    The first term converges to zero (see the proof of \Cref{stm:zero_kl}).
    We have also already seen $\distr{P}_{\rparam \given \sparam, \dataset} = \gauss (\mu , \Sigma)$ (see the discussion around \Cref{eq:full_reparam}).
    Using the formula for KL divergence between two multivariate normal distributions
    \begin{align*}
            \KL (\gauss(0, I_{\rdimParam}) \, \| \, \distr{P}_{\rparam \given \sparam, \dataset})
        &\propto
        \trace(\Sigma^{-1}) 
        +
        \| \mu \|_{\Sigma}^{-1}
        -
        \rdimParam
        +
        \log \det (\Sigma)
        \\
        &=
        \lstd^{-2}
        \trace(\enngp)
        -
        \lstd^{-2} [
            I_{\nPoints}
            -
            \lstd^{-2}
            \embnngp (
                I_{\rdimParam}
                +
                \lstd^{-2}
                \embnngp^\top \embnngp
            )^{-1}
            \embnngp^\top
            -
            I_{\nPoints}
        ]
        -
        \log \det (\lstd^{-2} \regnngp[\lstd^2])
        \\
        &=
        \lstd^{-2}
        \trace(\enngp)
        +
        \| y \|_{\lstd^{-2} I_\nPoints - \regnngp[\lstd^2]^{-1}}^2
        +
        \nPoints \log(\lstd^2)
        -
        \log \det (\regnngp[\lstd^2])
        \, ,
    \end{align*}
    where we used the Woodbury identity for the last equality.
    Convergence of all the above terms and their expectations has already been established in the proof of \Cref{stm:zero_kl}.
    Reintroducing the dropped constant, we therefore have
    \begin{align*}
        \KL ( \gauss(0, I_{\dimParam}) \, \| \, \posterior )
        \to
        \tfrac{1}{2} \bigl[
            \lstd^{-2}
            \trace(\nngp)
            +
            \| y \|_{\lstd^{-2} I_\nPoints - \nngp_{\lstd^2}^{-1}}^2
            +
            \nPoints \log(\lstd^2)
            -
            \log \det ( \nngp_{\lstd^2} )
        \bigr]
        \, .
    \end{align*}
    
    To establish equivalence to the KL between the NNGP prior and posterior, it is sufficient to employ equation~(12) from \citep{matthews2016kl} with, in their notation, $\distr{Q} = \distr{P}$ set to be the prior.
    This results in
    \begin{align*}
        \log Z
        +
        \tfrac{1}{2} \biggl[
            \nPoints \log (2\pi \sigma^2)
            +
        &    \lstd^{-2}
            \E_{\fn(X) \sim \gauss (0, \nngp)} \bigl[
                \| y - \fn(X) \|_2^2
            \bigr]
        \biggr]
        =
        \log Z 
        +
        \tfrac{1}{2} \biggl[
            \nPoints \log (2\pi \sigma^2)
            +
            \lstd^{-2}
            [
                \| y \|_2^2
                +
                \trace(\nngp)
            ]
        \biggr]
        \, ,
    \end{align*}
    where
    \begin{align*}
        \log Z
        &= 
        \E_{\fn(X) \sim \gauss(0, \nngp)} \bigl[
            (2\pi\lstd^2)^{-n/2} \exp \bigl\{
                -\tfrac{1}{2\lstd^2}
                \| y - \fn(X) \|_2^2
            \bigr\}
        \bigr]
        \\
        &=
        -\tfrac{1}{2} \bigl[
            \nPoints \log (2\pi\lstd^2)
            +
            \log \det (\nngp)
            +
            \log \det (\nngp^{-1} + \lstd^{-2} I_\nPoints)
            +
            \| y \|_{\nngp_{\lstd^2}^{-1}}^{2}
        \bigr]
        \\
        &=
        -\tfrac{1}{2} \bigl[
            \nPoints \log (2\pi)
            +
            \log \det (\nngp_{\lstd^2})
            +
            \| y \|_{\nngp_{\lstd^2}^{-1}}^{2}
        \bigr]
        \, ,
    \end{align*}
    by basic determinant identities.
    Substituting $\log Z$ into the previous equation yields the result.
\end{proof}

\subsection{\texorpdfstring{\Cref{stm:gradient_convergence}}{Density gradient convergence}}
\label{app:grad_convergence}

The two additional technical assumptions mentioned in the main text are:
\begin{enumerate}
    \item The set of last layer preactivations $\fn^{\nLayers}$ is exchangeable over the embedding (width) index, and converges in distribution to a well-defined limit (not necessarily a GP, even though that is the case in most scenarios).
    These facts are true for almost all architectures---\citep[see, e.g.,][]{matthews2018gaussian,alonso2018deep,hron2020infinite,yang2020tensor}.

    \item Defining $\Tilde{\ntk}_{11} \coloneqq \tfrac{\partial \hidden_{:1}}{\partial \spriorParam} \tfrac{\partial \hidden_{:1}^\top}{\partial \spriorParam} \in \R{\nPoints \times \nPoints}$, we assume $\E [ \| \Tilde{\ntk}_{11}  \|_2^2 ]$ converges as well.
    This will again be generally true by $\| \Tilde{\ntk}_{11}  \|_2^2 \leq \trace(\Tilde{\ntk})^2$ and theorem~A.2 in \citep{yang2020tensor}.
\end{enumerate}

\gradientConvergence*

\begin{proof}[Proof of \Cref{stm:gradient_convergence}]
    By $\dens{p}(\rpriorParam \given \spriorParam, \dataset) \sim \gauss(0, I_\rdimParam)$ (see the discussion after \Cref{eq:full_reparam}), the last layer gradients always exactly agree with that of the standard normal, i.e., $\nabla_{\rpriorParam} \log e^{- \| \rpriorParam \|_2^2 / 2} = - \rpriorParam$.
    Hence $\Delta_\priorParam$ depends only on $\spriorParam$.
    Substituting from \Cref{eq:remainder_posterior} (after application of the Weinstein--Aronszajn identity)
    \begin{align*}
        \| \Delta_\priorParam \|_2
        \leq
        \| \nabla_{\spriorParam} \log \det(\regnngp[\lstd^2])  \|_2
        +
        \| \nabla_{\spriorParam} y^\top \regnngp[\lstd^2]^{-1} y \|_2
        \, .
    \end{align*}
    We will use the shortcut $\partial_i$ to denote the derivative w.r.t.\ $\spriorParam_i$, so that $\| \nabla_{\spriorParam} \cdots \|_2 = \sum_{i=1}^{\sdimParam} (\partial_i \cdots)^2$.
    Applied to the individual summands, we have $\partial_i \log \det(\regnngp[\lstd^2]) = \trace (\regnngp[\lstd^2]^{-1} (\partial_i \enngp)) = 1_\nPoints^\top \regnngp[\lstd^2]^{-1} (\partial_i \enngp) 1_\nPoints$ where $1_\nPoints \in \R{\nPoints}$ is a vector of all ones.
    Similarly, $\partial_i \, y^\top \regnngp[\lstd^2]^{-1} y = - y^\top \regnngp[\lstd^2]^{-1} (\partial_i \enngp) \regnngp[\lstd^2]^{-1} y$.
    Hence the summands in both gradients have the form $( r^\top (\partial_i \enngp) s )^2$ for some possibly random vectors $r, s \in \R{\nPoints}$ whose distribution may depend on the width.
    Substituting the definition of $\embnngp$
    \begin{align*}
        \partial_i \enngp_{ab}
        \propto
        \frac{1}{\rdimParam}
        \sum_{j=1}^{\rdimParam}
            (\partial_i \hidden_{aj})
            \hidden_{bj}
            +
            \hidden_{aj}
            (\partial_i \hidden_{bj})
        \, ,
    \end{align*}
    where we used the abbreviation $\hidden_{aj} \coloneqq \hidden_j^\nLayers (x_a)$.
    Therefore
    $
        r^\top (\partial_i \enngp) s
        \propto
        \frac{1}{\rdimParam}
        \sum_j
            \langle r, \partial_i \hidden_{:j} \rangle
            \langle s, \hidden_{:j} \rangle
            +
            \langle s, \partial_i \hidden_{:j} \rangle
            \langle r, \hidden_{:j} \rangle
    $,
    so
    \begin{align*}
        \sum_{i=1}^{\sdimParam}
            (r^\top (\partial_i \enngp) s)^2
        &\propto
        \frac{1}{(\rdimParam)^2}
        \sum_{i=1}^{\sdimParam}
        \sum_{j, k=1}^{\rdimParam}
            [ r^\top (\partial_i \hidden_{:j}) (\partial_i \hidden_{:k})^\top r ]
            [ s^\top \hidden_{:j} h_{:k}^\top s ]
            +
            [ r^\top (\partial_i \hidden_{:j}) (\partial_i \hidden_{:k})^\top s ]
            [ s^\top \hidden_{:j} h_{:k}^\top r ] 
            + 
            \ldots
        \\
        &=
        \frac{1}{(\rdimParam)^2}
        \sum_{j, k=1}^{\rdimParam}
            [ s^\top \hidden_{:j} h_{:k}^\top s ]
            \bigl[ r^\top \tfrac{\partial \hidden_{:j}}{\partial \spriorParam} \tfrac{\partial \hidden_{:k}^\top}{\partial \spriorParam} r \bigr]
            +
            [ s^\top \hidden_{:j} h_{:k}^\top r ]
            \bigl[ r^\top \tfrac{\partial \hidden_{:j}}{\partial \spriorParam} \tfrac{\partial \hidden_{:k}^\top}{\partial \spriorParam} s \bigr]
            +
            \ldots
        \, ,
    \end{align*}
    where we have expanded the square (the last two terms are omitted as signified by the ellipsis).
    Defining $\Tilde{\ntk}_{jk} \coloneqq \tfrac{\partial \hidden_{:j}}{\partial \spriorParam} \tfrac{\partial \hidden_{:k}^\top}{\partial \spriorParam}$
    \begin{align*}
        \sum_{i=1}^{\sdimParam}
            (r^\top (\partial_i \enngp) s)^2
        \propto
        \frac{1}{(\rdimParam)^2}
        \sum_{j, k=1}^{\rdimParam}
            (\langle s, h_{:j} \rangle \, r + \langle r , h_{:j} \rangle \, s)^\top
            \Tilde{\ntk}_{jk}
            (\langle s, h_{:k} \rangle \, r + \langle r , h_{:k} \rangle \, s)
    \end{align*}
    
    By \Cref{stm:zero_kl}, we can now integrate w.r.t.\ to the prior $\gauss(0, I_{\sdimParam})$ instead of the posterior $\distr{P}_{\spriorParam \given \dataset}$, because by Pinsker's inequality $\TV{\gauss(0, I_{\sdimParam})}{\distr{P}_{\spriorParam \given \dataset}} \lesssim [\KL (\gauss(0, I_{\sdimParam}) \, \| \, \distr{P}_{\spriorParam \given \dataset})]^{1/2}$ where the right-hand side goes to zero as widths go to infinity (see the proof of \Cref{stm:zero_kl} for details).
    Using $z_j \coloneqq \langle s, h_{:j} \rangle \, r + \langle r , h_{:j} \rangle \, s \in \R{\nPoints}$
    \begin{align}\label{eq:gradient_norm_ntk}
        \sum_{i=1}^{\sdimParam}
            \E [(r^\top (\partial_i \enngp) s)^2 ]
            =
            \tfrac{1}{\rdimParam}
            \E [
                z_1^\top
                \Tilde{\ntk}_{11}
                z_1
            ]
            +
            (1 - \tfrac{1}{\rdimParam})
            \E [
                z_1^\top
                \Tilde{\ntk}_{12}
                z_2
            ]
    \end{align}
    by the assumed exchangeability of the readout units.
    By the linear readout assumption,
    $\Tilde{\ntk}_{ij}$ is just a scaled version of $\entk - \enngp$, where $\entk \coloneqq (\partial_\priorParam \fn^{\nLayers + 1} ) (\partial_\priorParam \fn^{\nLayers + 1} )^\top$ is the \emph{empirical NTK}, with the output layer `integrated out'.
    It thus converges in probability under the prior (and thus the posterior) by the above argument and \citep[theorem 2.10]{yang2020tensor}, with $\Tilde{\ntk}_{ij}$ going to a constant $\Tilde{\ntk} \in \R{\nPoints \times \nPoints}$ if $i = j$, and to zero otherwise.
    Since $\regnngp[\lstd^2]^{-1}$ also converges to a constant in probability (see the proof of \Cref{stm:zero_kl}), the vectors $r$ and $s$ do so as well.
    Since the vectors $h$ are just $\nonlin$ applied pointwise to the previous layer outputs, which converge to a GP by assumption, they will converge in distribution to the corresponding $\nonlin_\#$ pushforward by the continuity of $\nonlin$.
    Their limit will be denoted simply by $z$.
    By the Slutsky's lemma, the integrand $z_1^\top \Tilde{\ntk}_{11} z_1$ thus converges in distribution to $z^\top \Tilde{\ntk} z$, and $z_1^\top \Tilde{\ntk}_{12} z_2$ to zero.
    
    Since $|z_1^\top \Tilde{\ntk}_{12} z_2| \lesssim z_1^\top \Tilde{\ntk}_{11} z_1 + z_2^\top \Tilde{\ntk}_{11} z_2$ by the outer product definition of $\Tilde{\ntk}_{11}$, and the arithmetic-geometric mean inequality, it will be sufficient to establish uniform integrability (UI) of $z_1^\top \Tilde{\ntk}_{11} z_1$ to show the expectations converge \citep[theorem~3.5]{billingsley1999convergence}.
    This can be obtained by observing $z_1^\top \Tilde{\ntk}_{11} z_1 \leq \| \Tilde{\ntk}_{11} \|_2 \| z \|_2^2 \leq \| \Tilde{\ntk}_{11} \|_2^2 + \| z \|_2^4 \leq \trace(\Tilde{\ntk}_{11})^2 + \| z \|_2^4$ which are both non-negative random variables, which converge in probability (resp.\ in distribution) by continuity of the trace and the the square function.
    Their expectations then converge by theorem~A.2 in \citep{yang2020tensor}.
    The collection of $\{ z_1^\top \Tilde{\ntk}_{11} z_1 \}$ indexed by width is thus UI by \Cref{stm:ui_group_bound}.
    Hence the expectations on the right hand side of \Cref{eq:gradient_norm_ntk}
    converge, implying their sequence is bounded, and thus
    \begin{align*}
        \E [ \| \Delta_\priorParam \|_2 ]
        \leq
        \sqrt{\E [ \| \Delta_\priorParam \|_2^2 ]}
        \lesssim
        \tfrac{1}{\sqrt{\rdimParam}}
        +
        o(1)
        \, ,
    \end{align*}
    Convergence in posterior probability follows by the above Pinsker's inequality argument, and the Markov's inequality.
\end{proof}

\subsection*{Auxiliary results}

\begin{lemma}\label{stm:ui_group_bound}
    If $\mathcal{C}$ and $\mathcal{C}'$ are two collections of random variables such that for every $\rv{X} \in \mathcal{C}$ there exists $\rv{Y} \in \mathcal{C}'$ such that $|\rv{X}| \leq |\rv{Y}|$ a.s., and $\mathcal{C}'$ is uniformly integrable (UI), then $\mathcal{C}$ is uniformly integrable.
\end{lemma}
\begin{proof}[Proof of \Cref{stm:ui_group_bound}]
    Clearly $\sup_{\rv{X} \in \mathcal{C}} \E |\rv{X}| \leq \sup_{\rv{Y} \in \mathcal{C}'} \E |\rv{Y}| < \infty$ by UI.
    Furthermore, for arbitrary $\epsilon > 0$, $\exists \delta > 0$ s.t.\
    \begin{align*}
        \sup_{\Prob(A) \leq \delta}
        \sup_{\rv{X} \in \mathcal{C}}
            \E |\rv{X}| \indicator{A}
        \leq
        \sup_{\Prob(A) \leq \delta}
        \sup_{\rv{Y} \in \mathcal{C}'}
            \E |\rv{Y}| \indicator{A}
        \leq
        \epsilon
    \end{align*}
    again by the uniform integrability of $\mathcal{C}'$.
    Taken together, the two facts imply $\mathcal{C}$ is also uniformly integrable.
\end{proof}

\section{Feature-space vs.\ data-space reparametrisation}
\label{app:data_space_reparametrisation}

The \textbf{feature-space version} of our algorithm---presented in the main body and utilised in the experiments---scales cubically in the top layer width $\rdimParam$ rather than the number of datapoints $\nPoints$ (cubic scaling in $\nPoints$ is the main bottleneck of exact GP inference).
This is preferable in most practical circumstances since $\rdimParam$ is hardly ever significantly larger than $\nPoints$, making the additional computational overhead negligible \Cref{fig:speed}.
Here we present a \textbf{data-space version} which scales cubically in $\nPoints$ as in GPs, but only quadratically in $\rdimParam$.
This may be useful, e.g., when investigating properties of very wide BNNs.

The data-space version can be obtained using ideas related to the `kernel trick'.
Starting with $\posteriorMap(\priorParam) = \mu + \Sigma^{1/2} \priorParam$, we have $\mu = (\lstd^2 I_{\rdimParam} + \embnngp^\top \embnngp)^{-1} \embnngp^\top y = \embnngp^\top (\lstd^2 I_\nPoints + \embnngp \embnngp^\top )^{-1} y$.
For the covariance 
$$
\Sigma 
= (I_{\rdimParam} + \lstd^{-2} \embnngp^\top \embnngp)^{-1} 
= I_{\rdimParam} - \embnngp^\top (\lstd^2 I_\nPoints + \embnngp \embnngp^\top )^{-1} \embnngp 
\, ,$$
by the Woodbury identity.
The reparametrisation requires the square root $\Sigma^{1/2}$ though, for which we can use a \textbf{Woodbury-like identity for the matrix square root}.
To our knowledge, this identity is \emph{new} and may thus be of independent interest:
\begin{align}\label{eq:woodbury_sqrt}
    (I_p + A^\top A)^{1/2}
    =
    I_p
    + 
    A^\top \left[
        I_m + (I_m + A A^\top)^{1/2} 
    \right]^{-1}
    A
    \, ,
\end{align}
for any $m, p \in \natnum$ and $A \in \R{m \times p}$.
Clearly this reduces the $\mathcal{O}(p^3)$ complexity of the square root on the l.h.s.\ to the $\mathcal{O}(m^3 + m^3) = \mathcal{O}(m^3)$ complexity of the square root and matrix inverse on the r.h.s.
To see the identity holds, use SVD decomposition $A = USV^\top$ to rewrite the l.h.s.\ as $V (I_p + S^\top S)^{1/2} V^\top$, and the r.h.s.\ as 
$$
    I_p
    + 
    A^\top \left[
        I_m + (I_m + A A^\top)^{1/2} 
    \right]^{-1}
    A
    =
    V 
    \left\lbrace 
        I_p 
        +
        S^\top \left[
            I_m + (I_m + S S^\top)^{1/2}
        \right]^{-1} S 
    \right\rbrace 
    V^\top
    \, ,
$$
and observe that $\sqrt{1 + x^2} = 1 + \frac{x^2}{1 + \sqrt{1 + x^2}}$ for any $x \in \R{}$.
Since we need the inverse, $(I_{\rdimParam} + \lstd^{-2} \embnngp^\top \embnngp)^{-1/2} = \Sigma^{1/2}$, we can combine \Cref{eq:woodbury_sqrt}
with the \emph{usual} Woodbury identity for matrix inverse to obtain
\begin{align*}
    \Sigma^{1/2}
    =
    I_{\rdimParam}
    -
    \embnngp^\top \left[
        \sqrt{\lstd^2 I_\nPoints + \embnngp \embnngp^\top}
        \left(
            \lstd I_\nPoints
            +
            \sqrt{\lstd^2 I_\nPoints + \embnngp \embnngp^\top}
        \right)
    \right]^{-1}
    \embnngp
    \, .
\end{align*}
We can then perform the \emph{eigendecomposition} $\lstd^2 I_\nPoints + \embnngp \embnngp^\top = Q \Lambda Q^\top$ which has $\mathcal{O}(\nPoints^3)$ complexity (same as the Cholesky decomposition), and thus allows efficient computation of \emph{both} $\mu = \embnngp^\top Q \Lambda^{-1} Q^\top y$, and
\begin{align*}
    \Sigma^{1/2} \priorParam
    =
    \priorParam
    -
    \embnngp^\top
    Q \Lambda^{-1/2} (\lstd I_\nPoints + \Lambda^{1/2})^{-1} Q^\top
    \embnngp \priorParam
    \, ,
\end{align*}
since $\Lambda$ is a diagonal matrix.
Similar to \Cref{sect:cholesky}, we can \emph{reuse} the decomposition result to obtain the density value essentially for free using $\dens{p}(\priorParam \given \dataset) = \dens{p}(\param \given \dataset) \left| \det \partial_\priorParam \param \right|$, and \Cref{eq:determinant} together with the Weinstein--Aronszajn identity
\begin{align*}
    \left| \det \partial_\priorParam \param \right|
    =
    \sqrt{\det (\Sigma)}
    =
    \sqrt{\det (I_{\rdimParam} + \lstd^{-2} \embnngp^\top \embnngp)}
    \propto
    \sqrt{\det (\lstd^2 I_{\nPoints} + \embnngp \embnngp^\top)}
    =
    \prod_{i=1}^{\nPoints}
        \sqrt{\Lambda_{ii}}
    \, .
\end{align*}

\section{Extension to multi-dimensional outputs}
\label{app:multi_dim}

\newcommand{\dimOut}{\dimParam^{\nLayers + 1}}

\looseness=-1
In the main text, we assumed the output dimension $\dimOut \in \natnum$ is equal to one for simplicity of exposition.
The extension to $\dimOut > 1$ is simple.
In particular, we take the factorised Gaussian likelihood (see \Cref{app:non-gauss_likelihood} for the categorical likelihood)
$$
\log \dens{p}(y \given X, \param) 
=
- \frac{\nPoints \dimOut}{2} \log (2 \pi \lstd^2) 
- \frac{1}{2\lstd^2} \sum_{i=1}^n \sum_{j=1}^{\dimOut}
(y_{ij} - f_{\param}(x_i)_j)^2
\, ,
$$
where $f_\param(x_i) \in \R{\dimOut}$.
Each output dimension is associated with an independent readout weight $ \weight_{: j}^{\nLayers+1} \in \R{\rdimParam}$ s.t.\
$
f_{\param}(x)_j 
=
\nicefrac{\wstd^{\nLayers+1} \hidden(x)^\nLayers \weight_{: j}^{\nLayers+1}}{\sqrt{\rdimParam}}
$.
To compute $\param = \posteriorMap(\priorParam)$, the only modification occurs in the last layer where the mean $\mu_{: j} = \embnngp^\top (\lstd^2 I_\nPoints + \embnngp \embnngp^\top)^{-1} y_{: j}$ is different for each $ \weight_{:j}^{\nLayers+1}$, with $y_{: j} \in \R{\nPoints}$ the $j$\textsuperscript{th} column of the targets $y \in \R{\nPoints \times \dimOut}$.
Importantly, $\embnngp$ is the same for all $j$, meaning a single Cholesky decomposition is sufficient.
Modern solvers allow for multi-column r.h.s.\ meaning that we can solve for the reparametrised value of the output weights at once, limiting the computational overhead.

The only other change is in the determinant $\det (\partial_\priorParam \param)$.
Because each $ \weight_{:j}^{\nLayers+1}$ only depends on $\spriorParam = \sparam$, the determinant still has the upper-triangular structure analogous to \Cref{eq:determinant}, implying 
\begin{align*}
    \log \left| \det (\partial_\priorParam \param) \right|
    =
    \text{constant} 
    + 
    \dimOut \log \det( \lstd^2 I_\nPoints + \embnngp \embnngp^\top )
    \, .
\end{align*}
We can thus again re-use the result of the Cholesky decomposition to obtain the determinant value essentially for free.

\section{Repriorisation induced LMC speed-up in linear models}
\label{app:lmc_speedup_lin}

While the linear case $f_\param(x) = \langle x, \param \rangle$ is of course not equivalent to deep BNNs, we use it here to provide a rough intuition for the source and magnitude of the improvement that could be encountered when running LMC with $\dens{p}(\priorParam \given \dataset)$ instead of $\dens{p} (\param \given \dataset)$.
To sample from a generic density $\dens{p} (z)$,
LMC (and HMC) typically uses the Leapfrog integrator to simulate the Hamiltonian dynamics of a system with \emph{potential energy} $U(z) = - \log \dens{p}(z)$
and \emph{kinetic energy} $K(m) = \tfrac{1}{2} \| m \|_{M^{-1}}^2$, 
where $m \in \R{d}$ is an auxiliary momentum variable, and $M \in \R{\dimParam \times \dimParam}$ a positive definite mass matrix 
\citep{brooks2011handbook}.

The integrator simulates movement of a particle with initial state $(z_0, m_0) \in \R{\dimParam} \times \R{\dimParam}$ along a trajectory of constant energy $H(z, m) = U(z) + K(m)$ in the `counter-clockwise direction'
\begin{align*}
    \begin{pmatrix}
        \Dot{m}_t \\
        \Dot{z}_t
    \end{pmatrix}
    =
    \begin{pmatrix}
        O & - I_\dimParam \\
        I_\dimParam & 0
    \end{pmatrix}
    \begin{pmatrix}
        \partial_{m_t} H(z_t, m_t) \\
        \partial_{z_t} H(z_t, m_t)
    \end{pmatrix}
    =
    \begin{pmatrix}
        - \nabla \, U(z_t) \\
        \, \nabla K(m_t)
    \end{pmatrix}
    \, .
\end{align*}
A nice property of the linear case is that we can easily solve this matrix ODE, and thus determine the error accrued by the integrator for any given $t > 0$.\footnote{Whether this can be exploited to design a more efficient integrator for wide BNNs is an interesting question for future research.}
Without reparametrisation ($z = \param$), $\nabla U (\param) = \param + \lstd^{-2} X^\top (X \param - y) = C \theta - b_\dataset$, where
$C = I + \lstd^{-2} X^\top X = \Sigma^{-1}$ and $b_\dataset = \lstd^{-2} X^\top y = \Sigma^{-1} \mu$.
This translates into the matrix ODE
\begin{align*}
    \begin{pmatrix}
        \Dot{m}_t \\
        \Dot{\param}_t
    \end{pmatrix}
    =
    \underbrace{
        \begin{pmatrix}
            O & - C \\
            M^{-1} & 0
        \end{pmatrix}
    }_{= A}
    \begin{pmatrix}
        m_t \\
        \param_t
    \end{pmatrix}
    +
    \underbrace{
        \begin{pmatrix}
            b_\dataset \\
            0
        \end{pmatrix}
    }_{= A b}
    \, ,
\end{align*}
which has the usual solution
\begin{align*}
    \begin{pmatrix}
        m_t \\ 
        \param_t
    \end{pmatrix}
    =
    (I - e^{t A}) b
    +
    e^{t A}
    \begin{pmatrix}
        m_0 \\
        \phi_0
    \end{pmatrix}
    \, .
\end{align*}
Accuracy of the integrator is thus determined by how closely it can approximate the matrix exponential $e^{tA}$.
The value of this exponential is known; to simplify, we take $M = I$ as used in our experiments
\begin{align*}
    e^{t A}
    &=
    \sum_{k=0}^\infty 
        \frac{t^{2k}}{(2k)!}
        \begin{pmatrix}
            (- C)^k & 0 \\
            0 & (- C)^k
        \end{pmatrix}
    +
    \sum_{k=0}^\infty
        \frac{t^{2k + 1}}{(2k + 1)!}
        \begin{pmatrix}
            0 & (- C)^{k + 1} \\
            (- C)^k & 0
        \end{pmatrix}
    \\
    &=
    \begin{pmatrix}
        \cos(t \sqrt{C}) & - \sqrt{C} \sin(t \sqrt{C}) \\
       (\sqrt{C})^{-1} \sin(t \sqrt{C}) & \cos(t \sqrt{C} )
    \end{pmatrix}
    \, .
\end{align*}
Note that $C$ is the inverse posterior covariance, and the above equations describe a `rotation' along an ellipsoid defined by the eigendecomposition of $C$, as expected.
Most importantly, the integrator error will be driven by $t \sqrt{C}$, where in particular the LMC \emph{stepsize has to be inversely proportional to $\| \sqrt{C} \|_2$} if a local approximation of $\sin$ and $\cos$ is to be accurate.

\looseness=-1
An analogous argument holds for the reparametrised version ($z = \priorParam$), where $\nabla U (\priorParam) = \smash{\tfrac{\partial \param}{\partial \priorParam}^\top} [\Sigma^{-1} (\lstd^{-2} \Sigma X^\top y + \sqrt{\Sigma} \, \priorParam) - \lstd^{-2} X^\top y] = \priorParam$ by $\Sigma = C^{-1}$ and $\tfrac{\partial \param}{\partial \priorParam} = \sqrt{\Sigma}$.
The stepsize can therefore be (inversely) proportional to $\| I_\dimParam \|_2 = 1$, disregards of both $\nPoints$ and $\dimParam$.
In contrast, without the reparametrisation, the stepsize must be \emph{inversely proportional to} (see above)
$$
\| \sqrt{C} \|_2 
=
\sqrt{\| C \|_2}
= 
\sqrt{1 + \tfrac{\|X^\top X\|_2}{\lstd^{2}}} 
\sim
\frac{\sqrt{n}}{\lstd}
\, ,
$$
where we used $\| X^\top X \|_2 = \nPoints \, \| \tfrac{1}{\nPoints} \sum_{i} x_i x_i^\top \|_2$ and assumed $\| \tfrac{1}{\nPoints} \sum_{i} x_i x_i^\top \|_2 =\mathcal{O}(1)$.
This assumption will be true for most sufficiently well-behaved distributions over $x_i$ by the law of large numbers (if $\E [xx^\top]$ exists), and the continuity of the operator norm \citep[see chapter~6 in][for a more detailed discussion]{wainwright2019high}.

\looseness=-1
In summary: without the reparametrisation, the maximum stepsize scales as $\tfrac{\lstd}{\sqrt{\nPoints}}$, whereas it is independent of $\nPoints$ and $\lstd$ in the reparametrised case.
The higher the $\nPoints$ and/or lower the $\lstd$, the bigger the advantage of our reparametrisation.
In the deep BNN case, we replace $X$ with $\embnngp$, and look at the eigenspectrum of the NNGP kernel divided by $\nPoints$ (which is related to its kernel integral operator in wide enough BNNs).
While only an approximation in deep BNNs, we note that a constraint of the above form must be satisfied by the sampler in order to successfully sample the readout layer (with the asymptotic NNGP replaced by the empirical one), which is why we intuitively expect the stepsize to scale \emph{at least} as $\tfrac{\lstd}{\sqrt{\nPoints}}$.
This was observed in our experiments, where we indeed applied this stepsize scaling for the final hyperparameter sweep used in \Cref{fig:ess_width_dataset,fig:ess_three_subsets}.

\section{Alternative reparametrisations}
\label{app:alt_reparam}

\subsection{Reparametrising other layers than readout}
\label{app:prereadout_reparam}

The reparametrisation $\posteriorMap \colon \priorParam \mapsto \param$ we define in \Cref{eq:full_reparam} ensures that $\rparam$ follows its true conditional posterior when $\rpriorParam \sim \gauss (0 , I_{\rdimParam})$, which we have seen is the case under $\dens{p}(\priorParam \given \dataset)$.
One may wonder whether there exists an alternative reparametrisation(s) $\altReparam \colon \priorParam \mapsto \theta$ which still ensures the KL divergence between $\dens{p}(\altReparam(\priorParam) \given \dataset) | \! \det \partial_\priorParam \altReparam(\priorParam)|$
and $\gauss(0, I_\dimParam)$ goes to zero as $\minWidth \to \infty$, 
but maps some other layer than the readout to its conditional posterior.
In other words, is the readout `special', or is mapping of \emph{any} one layer to its conditional sufficient to ensure reversion to the prior in the wide limit?

For an example of a case where reparametrisation of a layer other than the readout is sufficient, consider the \emph{linear} 1-hidden layer FCN with a \emph{single} input and output dimension, $\fn(x) = (x \cdot u^\top) v / \sqrt{\dimParam^{1}}$, with $u, v \in \R{\dimParam^1}$ respectively the input and readout weights (w.l.o.g.\ $\wstd = 1$).
By symmetry, $f(x) = (x \cdot v^\top) u / \sqrt{\dimParam^{1}}$, implying we can apply the $\posteriorMap$ from \Cref{eq:full_reparam} to the \emph{input} weights $u$, and use \Cref{stm:zero_kl} (mutatis mutandis) to obtain the desired KL-convergence.

\looseness=-1
However, this argument no longer holds when we introduce a nonlinearity, as the symmetry between $u$ and $v$ is broken.
Moreover, we now show that the KL-convergence to the prior does not happen if we consider the only available alternative $\altReparam$, i.e., mapping $u$ to its conditional posterior given the readout $v$ and $\dataset$,\footnote{A deterministic map from prior to posterior will exist because we are in $\R{\dimParam^1}$, and both prior and posterior are continuous.} and leaving the \emph{readout} $v$ without reparametrisation.

To start, observe that the conditional distribution of $v$ given $u$ and $\dataset$ remains
$\gauss( \mu , \Sigma )$ with $\mu = \embnngp^\top \regnngp[\lstd^2]^{-1} y$ (kernel trick), and $\smash{\Sigma = I - \embnngp^\top \regnngp[\lstd^2]^{-1} \embnngp}$ (Woodbury identity).
For simplicity, we prove that KL-convergence to prior cannot happen when the nonlinearity $\nonlin$ is \emph{bounded}.
Using a proof-by-contradiction, assume the KL \emph{does} converge to zero. 
By the chain rule for KL divergence, this implies the KL between the \emph{reparametrised marginal posterior} of the readout $v$ given $\dataset$ converges in KL to the prior over $v$, i.e., $\gauss(0, I_{\dimParam^1})$.
By the assumed boundedness of $\nonlin$ and the Pinsker's inequality, this in turn implies
\begin{align*}
    \E [v_{:1} \given \dataset]
    =
    \E [\embnngp_{:1}^\top \regnngp[\lstd^2]^{-1} y \given \dataset] 
    \to 
    \E_{\varepsilon \sim \gauss(0, I_{\dimParam^0})} [\nonlin(X \varepsilon)]^\top \nngp_{\lstd^2}^{-1} y
    \, ,
    \quad
    \text{as }
    \minWidth \to \infty
    \, .
\end{align*}
However, the Pinsker's inequality and boundedness of $\nonlin$ also imply that $\E[v_{:1} \given \dataset ] \to \E[ v_{:1} ] = 0$ (prior mean), which is a contradiction unless $\E_{\varepsilon \sim \gauss(0, I_{\dimParam^0})} [\nonlin(X \varepsilon)] = 0$
(or $y = 0$).
This is not true, e.g., for the popular sigmoid non-linearity.

\looseness=-1
In summary, the answer to our question from the first paragraph is `it depends'.
There are some linear networks in which symmetry enables reparametrising layers other than the readout and still obtaining the KL-convergence.
However, introduction of non-linearities breaks this symmetry, which in turn prevents reversion to the prior.
Reparametrisation of the readout is thus indeed `special', as it is the only layer which ensures convergence in \emph{all} cases (modulo the assumptions of \Cref{stm:zero_kl}).


\begin{figure*}[htbp]
    \centering
    \includegraphics[keepaspectratio,width=0.9\textwidth]{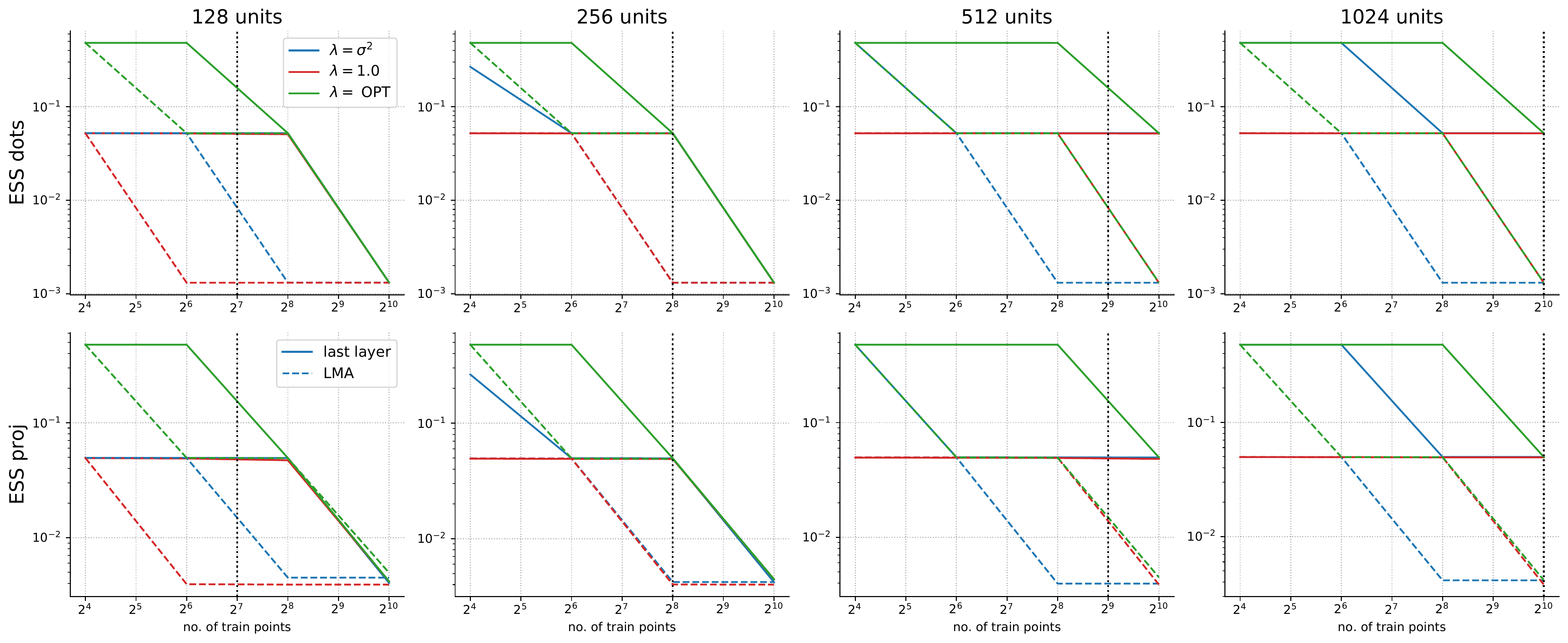}
    \caption{
    \looseness=-1
    \textbf{Comparing NNGP $\posteriorMap$ and NTK $\altReparam$} in terms of
    ESS per-step in the $\param$ space for subsets of \texttt{cifar-10}.
    \texttt{ESS proj} is defined the same as in \Cref{sect:experiments} (showing only the mean over projection dimensions).
    \texttt{ESS dots} is the ESS of the squared norm statistic $\| \param \|_2^2$.
    Since these are older experiments, $\posteriorMap$ is described as `last layer' (solid lines), and $\altReparam$ as `LMA' (dashed lines).
    Colours distinguish different values of the regulariser $\lambda$. 
    The $\lambda = \text{OPT}$ corresponds to the maximum ESS per-step values optimised over 
    $
    \log \lambda 
    = -4, -3, \ldots , 1
    $
    for each dataset size separately.
    While the NTK reparametrisation is sometimes competitive, the NNGP one is clearly preferable.
    }
    \label{fig:ntk_vs_nngp_reparam}
\end{figure*}

\subsection{NTK reparametrisation}
\label{app:ntk_reparam}

Another alternative is to \emph{linearise} $\fn$, i.e., to use the network Jacobian $J \coloneqq \partial_\priorParam \fn_{\priorParam} (X)$ to define the reparametrisation
\begin{align}\label{eq:lma}
    \param 
    = 
    \altReparam(\priorParam)
    =
    \priorParam 
    + 
    J^\top \regntk^{-1} (y - \fn_{\priorParam} (X))
    \, , \quad
    \text{with } 
    \regntk
    \coloneqq
    \lambda I_\dimParam
    +
    \entk
    =
    \lambda I_\dimParam
    +
    J J^\top
    \, , \,
    \text{for some }
    \lambda \geq 0
    \, .
\end{align}
This reparametrisation is inspired by the NTK theory where optimising the NN linearised around its (random) initial point by gradient descent yields the same large width behaviour as optimising the original NN \citep{lee2019wide}.
\emph{Informally}, if $\priorParam \sim \gauss (0, I)$ (the initialisation distribution in the NTK world), and the NN is wide enough
\begin{gather*}
    f_{\altReparam(\priorParam)}(x)
    \approx
    f_\priorParam(x)
    +
    J_x (\altReparam(\priorParam) - \priorParam)
    =
    f_\priorParam(x)
    +
    J_x J^\top \regntk^{-1} (y - \fn_{\priorParam} (X))
    \, , 
\end{gather*}
where $J_x \coloneqq \partial_\priorParam f_{\priorParam}(x)$.
By the NNGP theory, $f_\priorParam \sim \gp (0, k)$ is the NNGP limit;
by the NTK theory, $\entk_{x X} \coloneqq J_x J^\top \to \ntk_{x X} \in \R{1 \times \nPoints}$ and $\entk \to \ntk \in \R{\nPoints \times \nPoints}$ in probability as $\minWidth \to \infty$, where $\ntk$ is the NTK \citep[see, e.g.,][for an overview]{yang2020tensor}.
If we approximate by substituting the limits, $f_{\altReparam(\priorParam)} \approx f_\priorParam + \ntk_{\cdot X} \ntk_{\lambda}^{-1} (y - f_{\priorParam}(X))$,
we notice this is a fixed \emph{linear} transformation of the NNGP, and thus also a GP.
In fact, it is the NTK-GP when $\lambda = 0$ \citep[corollary~1]{lee2019wide}, i.e., the distributional limit of NNs optimised by gradient descent, where the randomness comes from the initialisation $\gauss(0, I)$.

The above sketch shows that $\param = \altReparam(\priorParam)$ provides parameter-space samples which converge to the NTK-GP limit when $\priorParam \sim \gauss(0, I)$, forming an \emph{informal} NTK counterpart to \Cref{stm:kl_functions}.
An alternative is to interpret \Cref{eq:lma} as a \emph{single} step of the Levenberg--Marquardt algorithm \citep[LMA;][]{levenberg1944method,marquardt1963algorithm}.
This perspective inspired us to also experiment with \emph{multi}-step LMA update (i.e., iterative application of $\altReparam$ from \Cref{eq:lma}) so as to adjust for potential inaccuracy of the linearisation;
we also experimented with just using multiple steps of \emph{gradient descent} (GD), i.e.,
iterative application of $\priorParam \mapsto \priorParam + \alpha J^\top (y - f_\priorParam(X))$ with a fixed $\alpha$.
Neither the GD update nor the iterative application worked significantly better than \Cref{eq:lma} in our experiments, which is why we abandon their further discussion here.

\looseness=-1
In the remainder of this section, we discuss the technical details of the NTK reparametrisation (\Cref{eq:lma}) implementation.
\Cref{fig:ntk_vs_nngp_reparam} shows a comparison of this $\altReparam$ with the `NNGP reparametrisation' $\posteriorMap$ we presented in the main paper (\Cref{eq:full_reparam}).
While $\altReparam$ is sometimes competitive with $\posteriorMap$, $\posteriorMap$ is clearly better overall.
This is likely because $\altReparam$ yields samples from the NTK-GP, which may induce a complicated $\dens{p}(\priorParam \given \dataset)$ when the NTK-GP significantly differs from the NNGP (i.e., the correct posterior, \citealp{hron2020exact}).
Since $\posteriorMap$ is also orders of magnitude more computationally efficient (scales with the top layer embedding size $\rdimParam$ instead of the full parameter space dimension $\dimParam$), and we have rigorous understanding of its behaviour (\Cref{sect:reparam}), we do not recommend using the NTK reparametrisation.
The motivation for describing it here is to inform any potential future research, 
documenting what we tried, and contrasting it with the NNGP reparametrisation.






\subsubsection{Computing \texorpdfstring{$\altReparam(\priorParam)$}{the NTK parameter value}}

Denoting the residuals by $\residual \coloneqq y - \fn_{\priorParam} (X)$, $\altReparam$ becomes
$\priorParam \mapsto \priorParam + J^\top \regntk^{-1} \residual$.
Because computing $\residual$ costs about the same as the forward pass, and multiplication by $J^\top$ the same as the backward pass (vector-Jacobian product), the only operations not common in standard NN training are the computation of $\regntk$, and the subsequent application of the linear solver $\residual \mapsto \regntk^{-1} \residual$.

\looseness=-1
Fortunately, the matrix $\entk = J J^\top$, known as the \emph{empirical NTK}, has recently become a subject of significant interest.
We can thus use the \emph{Neural Tangents} library \citep{neuraltangents2020} which provides a very efficient way of computing 
$\entk$,
and related matrix-vector products \citep{novak2021fast}. 
As for the $\residual \mapsto \regntk^{-1} \residual$, we tried several implicit and explicit solvers, approximate and exact, but the fastest and most numerically stable was the explicit Cholesky solver.
We recommend using at least \texttt{float32} precision (or its emulation if on TPU) to prevent significant deterioration of the acceptance probability.

Another possible approximation is subsetting the data.
Specifically, we can replace the $r$, $J^\top$, and $JJ^\top$ terms in $\param = \priorParam + J^\top (\lstd^2 I_\nPoints + JJ^\top)^{-1} r$ by their equivalent evaluated on a fixed data subset.
This may be especially tempting when computational and memory bottlenecks are an issue.
One nevertheless has to be careful about designing any approximations to $\altReparam$ as---especially in the data subsetting case---these may inadvertently cause the resulting $\dens{p} (\priorParam \given \dataset)$ to be as constrained as $\dens{p}(\param \given \dataset)$, and thus erase any possible LMC speed up.
For example, in the linear case, if we subset the data and some of the omitted inputs is not close to the linear span of the included ones, there will be a posterior covariance eigenvalue as low as $\smash{\propto \tfrac{\lstd^2}{\nPoints}}$ (instead of $\propto 1$), which would necessitate scaling the LMC stepsize by $\smash{\tfrac{\lstd}{\sqrt{\nPoints}}}$ (cf.\ \Cref{app:lmc_speedup_lin}).

\looseness=-1
While we aim to approximate the $\altReparam$ (\Cref{eq:lma}), any approximation of $\altReparam$ \emph{itself} is a valid reparametrisation (if bijective and differentiable), as long as the gradients and $\left| \det \partial_\priorParam \param \right|$ are both computed for the \emph{approximated} instead of the true $\altReparam$.

\subsubsection{Approximating \texorpdfstring{$\det \partial_\priorParam \altReparam(\priorParam)$}{the determinant}}

Unlike above, the approximations her \emph{can} affect the stationary distribution of the Markov chain or even prevent its existence.
Hence the obtained samples may not be drawn from the true posterior even in the infinite \emph{time} limit.

\looseness=-1
Our log determinant approximation combines the Taylor expansion of $z \mapsto \log (1 + z)$ around $z = 0$, and the Hutchinson trick for trace estimation.
In particular, for a real \emph{symmetric} matrix $A$ with $\| A \|_2 < 1$
\begin{align}\label{eq:logdet_taylor}
    \log |\! \det (I + A) |
    =
    -
    \sum_{j=1}^\infty
        \frac{(-1)^j}{j}
        \trace (A^j)
    \approx
    -
    \frac{1}{S}
    \sum_{i=1}^S
    \sum_{j=1}^M
        \frac{(-1)^j}{j}
        s_i^\top A^j s_i
    \, ,
\end{align}
where $s_i$ are i.i.d.\ zero mean vectors with $E[s_i s_i^\top] = I$.
We use vectors with i.i.d.\ Rademacher entries 
throughout.\footnote{We tried several alternatives to 
including Chebyshev and learned polynomials,
but did not observe a significant difference.
}

Unfortunately, substituting $A = \tfrac{\partial \param}{\partial \priorParam} - I$ into \Cref{eq:logdet_taylor} would violate the symmetry requirement since
\begin{align}\label{eq:lma_jacob}
    \frac{\partial \param}{\partial \priorParam}
    =
    (
        I 
        -
        J^\top
        \regntk^{-1}
        J
    )
    (I + H_{\weightedResidual})
    -
    J^\top
    \regntk^{-1}
    \left(
    \begin{smallmatrix}
        \text{---} & H_1 \, J^\top \weightedResidual & \text{---} \\ 
        & \vdots & \\
        \text{---} & H_\nPoints \, J^\top \weightedResidual & \text{---} \\
    \end{smallmatrix}
    \right)
    \, , 
    \quad
    \text{with }
    H_{\weightedResidual}
    \coloneqq
    \sum_{i=1}^\nPoints
        \weightedResidual_i
        H_i
    \, ,
\end{align}
where $\weightedResidual \coloneqq \regntk^{-1} r = \regntk^{-1} (y - \fn_\priorParam (X))$ is the `preconditioned residual', and 
$H_i \coloneqq \frac{\partial^2 \fn_\priorParam (x_i)}{\partial \priorParam^2}$ 
the Hessian at $x_i$.
A straightforward solution would be to use the identity $\smash{|\!\det(A)| = \sqrt{\det (AA^\top)}}$.
The downside is doubling of the computational time.
We thus instead approximate by
\begin{align}\label{eq:logdet_approx}
    \log \, \bigl|
        \det \bigl( 
            \tfrac{\partial \param}{\partial \priorParam} 
        \bigr)
    \bigr|
    \approx
    \text{constant}
    +
    \log |
        \! \det (
            I_\dimParam
            +
            H_{\weightedResidual}
        )
    |
    \, ,
\end{align}
Roughly speaking, this approximation exploits the fact that the column spaces of $I_\dimParam - J^\top \regntk^{-1} J$ and $J^\top$ in \Cref{eq:lma_jacob} are close to orthogonal for small $\lambda$ relative to $\smash{\| \entk \|_2}$.
(Careful when tuning $\lambda$!)
Rotating the space and using the block-matrix determinant formula allows us to decompose the log determinant into an asymptotically \emph{constant} component aligned with the column space of $J^\top$, and the \emph{stochastic} component $\log |\! \det (I_d +  H_{\weightedResidual})|$.

Combining \Cref{eq:logdet_taylor,eq:logdet_approx}, we have $A = H_{\weightedResidual}$ which is a symmetric matrix as desired.
Furthermore, the Hutchinson estimator only requires evaluation of the vector products $\smash{v \mapsto H_{\weightedResidual}^j v}$ which can be implemented efficiently by applying the JAX vector-Jacobian function to the map $\priorParam \mapsto J^\top \sg [\regntk^{-1} \residual]$ where $\sg$ is the stop-gradient function.
Finally, LMC requires estimation of the gradient but naive application of reverse-mode automatic differentiation may result in an \emph{out-of-memory} (OOM) error, especially for larger architectures (the Rademacher vectors $s$ have dimension $\dimParam$).
A more memory-efficient method is to compute both the value and gradient of the log determinant approximation in the forward pass.\footnote{\textbf{Caveat:} JAX allows only defining a custom $\mathtt{jvp}$, and can infer the $\mathtt{vjp}$ rule automatically. However, we found it more memory efficient to explicitly define the forward pass operation, cache the computed gradient, and use the cached value to evaluate the gradient.
}

\looseness=-1
To illustrate, consider the \emph{second-order Taylor expansion} ($M=2$ in \Cref{eq:logdet_taylor}) which we used in preliminary experiments.
Exploiting the symmetry of $A = H_{\weightedResidual}$, we define $\Tilde{s}_i \coloneqq A s_i$, and compute \emph{both} the log determinant and its gradient as
\begin{gather}\label{eq:val_grad_order2_general}
\begin{gathered}[c]
    \log \, \bigl|\det 
        \partial_\priorParam \altReparam(\priorParam)
    \bigr|
    \approx
    \frac{1}{S}
    \sum_{i=1}^S
        s_i^\top A s_i
        -
        \tfrac{1}{2}
        s_i^\top A^2 s_i
    =
    \frac{1}{S}
    \sum_{i=1}^S
            \Tilde{s}_i^\top
            (
                s_i
                -
                \tfrac{1}{2}
                \Tilde{s}_i
            )
    \, ,
    \\
    \nabla_\phi \log \, \bigl|\det 
        \partial_\priorParam \altReparam(\priorParam)
    \bigr|
    \approx
    \nabla_\priorParam
    \frac{1}{S}
    \sum_{i=1}^S
        s_i^\top A s_i
        -
        \tfrac{1}{2}
        s_i^\top A^2 s_i
    =
    \frac{1}{S}
    \sum_{i=1}^S
        \bigl(
            \tfrac{\partial \Tilde{s}_i}{\partial \priorParam}
        \bigr)^\top
        (s_i - \Tilde{s}_i)
    \, ,
\end{gathered}
\end{gather}
where $\smash{s_i - \Tilde{s}_i \mapsto ( \frac{\partial \Tilde{s}_i}{\partial \priorParam} )^\top (s_i - \Tilde{s}_i)}$ can be obtained via JAX $\mathtt{vjp}$.
Since evaluating all the summands at the same time leads to OOM issues,
we compute 
the result sequentially in a loop over mini-batches of a size smaller than $S$.\footnote{The first-order terms can be ignored for nonlinearities with zero second-derivative like ReLU as $\diag(H_i)=0$ here.}

\section{Non-Gaussian likelihoods}
\label{app:non-gauss_likelihood}

\begin{figure*}[htbp]
    \centering
    \includegraphics[keepaspectratio,width=0.5\textwidth]{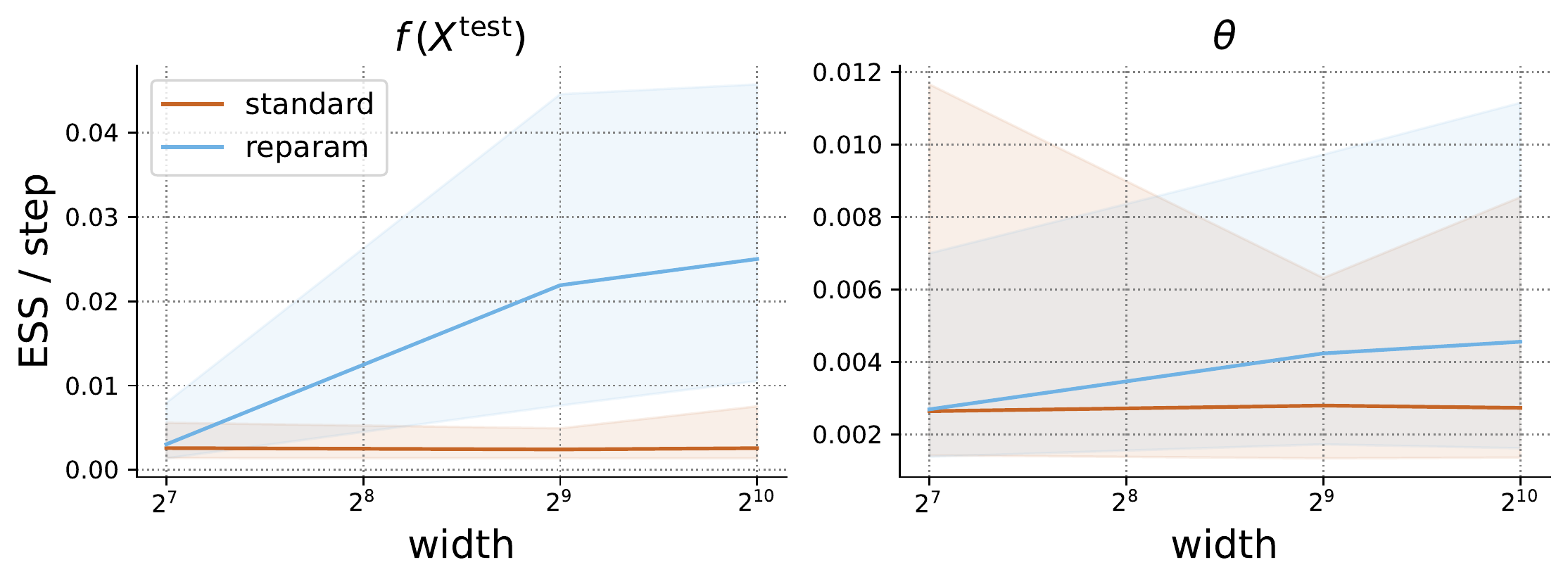}
    \caption{
        \textbf{Our reparametrisation can be combined with non-Gaussian likelihoods.}
        We ran a smaller version of the experiment in \Cref{fig:ess_width_dataset} (left) with the categorical likelihood, (much) less extensive hyperparameter tuning, and \emph{only} \texttt{100K} sampler steps.
        We had to lower the acceptance probability cutoff to 90\% (from 98\%) since the lowest stepsize in our grid was too high for the standard parametrisation (the reparametrised version achieved $\sim$99\% with a higher stepsize).
        The observed boost is thus likely an underestimate of the real gap.
        Mapping readout to its true conditional posterior induced by the categorical likelihood would likely work even better.
    }
    \label{fig:non-gauss_likelihood}
\end{figure*}

\end{document}